\documentclass[10pt,twocolumn,letterpaper]{article}

\usepackage[pagenumbers]{cvpr} 

\definecolor{cvprblue}{rgb}{0.21,0.49,0.74}
\usepackage[pagebackref,breaklinks,colorlinks,allcolors=cvprblue]{hyperref}

\usepackage{multirow}
\usepackage{listings}
\usepackage{soul}
\usepackage{dsfont}
\usepackage{amssymb}
\usepackage[misc]{ifsym}
\usepackage{tikz}
\usepackage[accsupp]{axessibility}  

\def\paperID{1525} 
\def\confName{CVPR}
\def\confYear{2025}

\title{DPFlow: Adaptive Optical Flow Estimation with a Dual-Pyramid Framework}

\author{Henrique Morimitsu\textsuperscript{1} \quad Xiaobin Zhu\textsuperscript{1 (\Letter)} \quad Roberto M. Cesar-Jr.\textsuperscript{2} \quad Xiangyang Ji\textsuperscript{3} \quad Xu-Cheng Yin\textsuperscript{1}\\
\textsuperscript{1}University of Science and Technology Beijing \quad \textsuperscript{2}University of S\~ ao Paulo \quad \textsuperscript{3}Tsinghua University\\
{\tt\small{ \{hmori, zhuxiaobin, xuchengyin\}@ustb.edu.cn \qquad rmcesar@usp.br \qquad xyji@tsinghua.edu.cn}}
}

\newcommand\copyrighttext{%
  \scriptsize \textcopyright 2025 IEEE. Personal use of this material is permitted. Permission from IEEE must be obtained for all other uses, in any current or future media, including reprinting/republishing this material for advertising or promotional purposes, creating new collective works, for resale or redistribution to servers or lists, or reuse of any copyrighted component of this work in other works.\\
  H. Morimitsu, X. Zhu, R. M. Cesar-Jr, X. Ji, and X.-C. Yin, “DPFlow: Adaptive Optical Flow Estimation with a Dual-Pyramid Framework” in CVPR, 2025. [Online]. DOI: 10.1109/CVPR52734.2025.01659. Available: https://ieeexplore.ieee.org/document/11094780}
\newcommand\copyrightnotice{%
\begin{tikzpicture}[remember picture,overlay]
\node[anchor=south,yshift=8pt] at (current page.south) {\fbox{\parbox{\dimexpr\textwidth-\fboxsep-\fboxrule\relax}{\copyrighttext}}};
\end{tikzpicture}%
}

\begin{document}\twocolumn[{%
\renewcommand\twocolumn[1][]{#1}%
\maketitle

\begin{center}
    \centering
    \captionsetup{type=figure}
    \begin{subfigure}{0.55\textwidth}
        \centering
        \includegraphics[width=\linewidth]{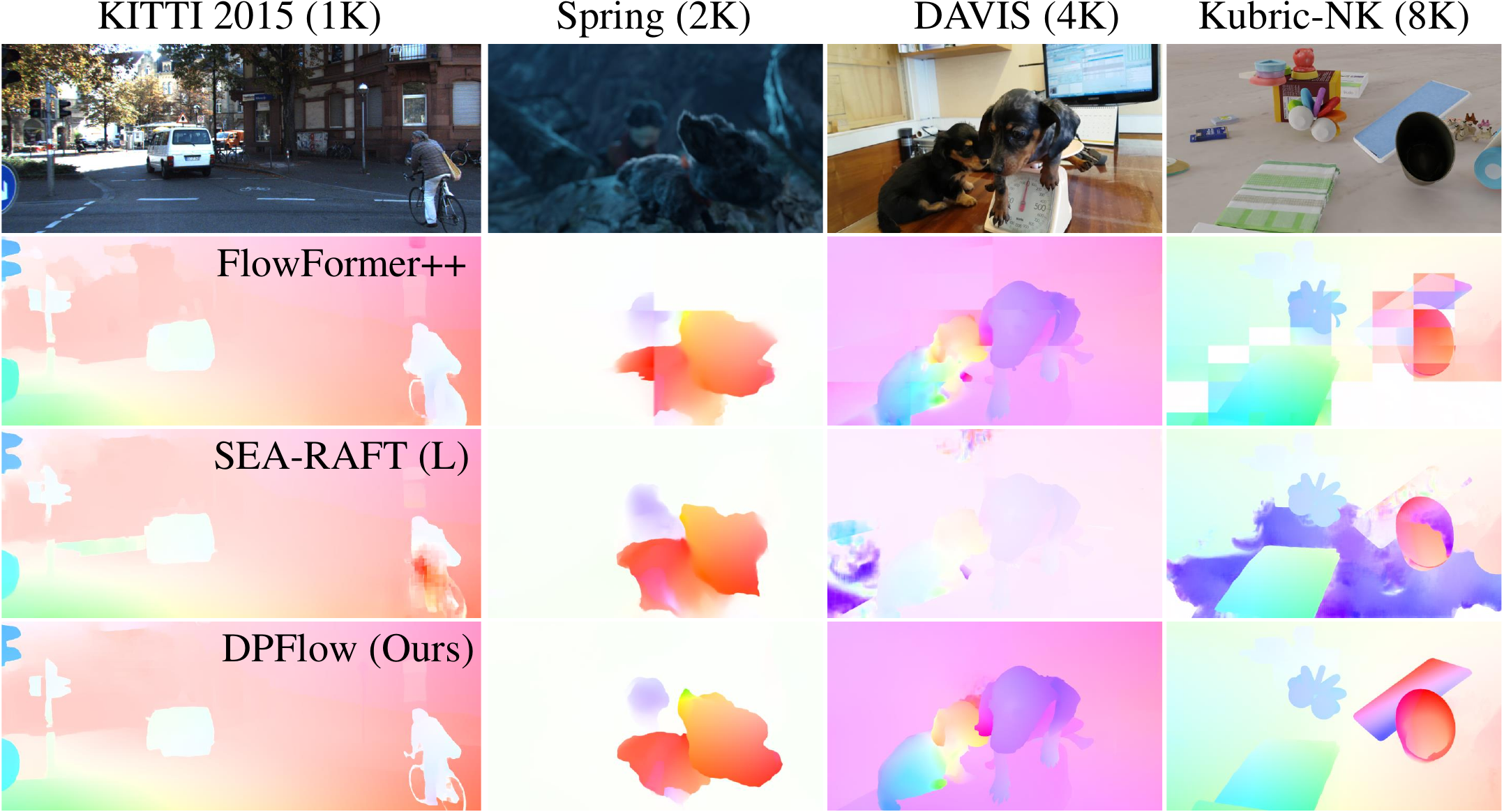}
        \caption{Qualitative results at 1K, 2K, 4K, and 8K resolutions.}
        \label{fig:teaser_qualitative}
    \end{subfigure}
    \qquad
    \begin{subfigure}{0.4\textwidth}
        \centering
        \includegraphics[width=\linewidth]{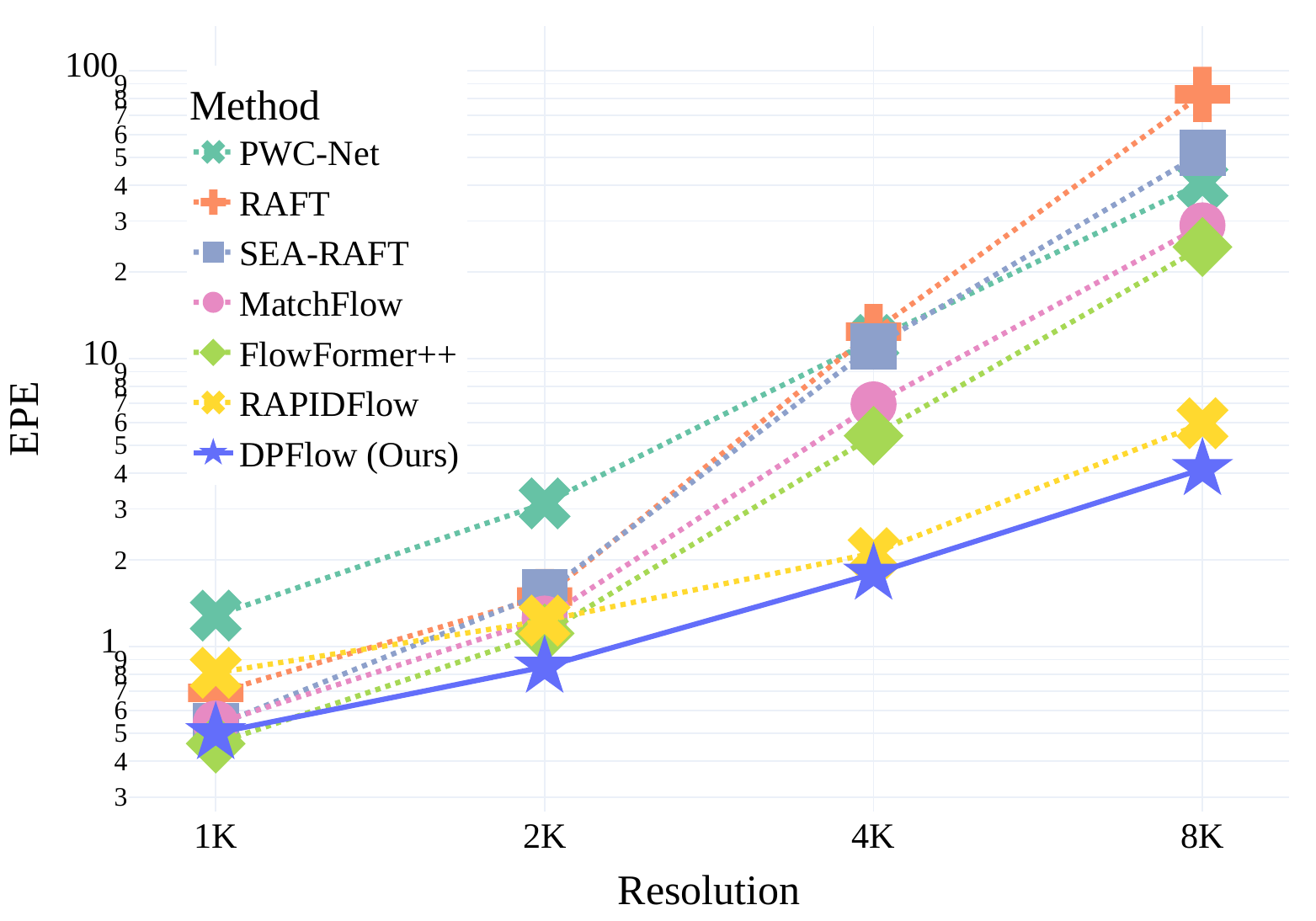}
        \caption{Results on the proposed Kubric-NK evaluation.}
        \label{fig:teaser_plot}
    \end{subfigure}
    \caption{(a) Optical flow results on multiple resolution images from the KITTI 2015~\cite{Menze2015Objectsceneflow}, Spring~\cite{Mehl2023SpringHighResolution}, DAVIS~\cite{PontTuset20172017DAVISChallenge} and our proposed Kubric-NK datasets. Methods like FlowFormer++~\cite{Shi2023FlowFormerMaskedCost} adopt tiling to handle high-resolution inputs, which causes square-shaped artifacts and loss of global context. Static approaches, such as SEA-RAFT~\cite{Wang2024SEARAFTSimple}, do not generalize well to large changes in resolution either. Our proposed DPFlow employs a flexible dual-pyramid design to adapt to larger inputs. (b) Our proposed Kubric-NK evaluation can quantify the generalization capabilities to resolutions up to 8K. At 8K, DPFlow outperforms RAPIDFlow~\cite{Morimitsu2024RAPIDFlowRecurrentAdaptable} by $30\%$ and FlowFormer++ by $83\%$.}
    \label{fig:teaser}
\end{center}%
}]

\copyrightnotice

\begin{abstract}
Optical flow estimation is essential for video processing tasks, such as restoration and action recognition.
The quality of videos is constantly increasing, with current standards reaching 8K resolution.
However, optical flow methods are usually designed for low resolution and do not generalize to large inputs due to their rigid architectures.
They adopt downscaling or input tiling to reduce the input size, causing a loss of details and global information.
There is also a lack of optical flow benchmarks to judge the actual performance of existing methods on high-resolution samples.
Previous works only conducted qualitative high-resolution evaluations on hand-picked samples.
This paper fills this gap in optical flow estimation in two ways.
We propose DPFlow, an adaptive optical flow architecture capable of generalizing up to 8K resolution inputs while trained with only low-resolution samples.
We also introduce Kubric-NK, a new benchmark for evaluating optical flow methods with input resolutions ranging from 1K to 8K.
Our high-resolution evaluation pushes the boundaries of existing methods and reveals new insights about their generalization capabilities.
Extensive experimental results show that DPFlow achieves state-of-the-art results on the MPI-Sintel, KITTI 2015, Spring, and other high-resolution benchmarks.
The code and dataset are available at \url{https://github.com/hmorimitsu/ptlflow/tree/main/ptlflow/models/dpflow}.
\end{abstract}
\section{Introduction}
\label{sec:intro}

Optical flow provides per-pixel video motion information, being essential for several video processing tasks, such as restoration~\cite{Lin2022UnsupervisedFlowAligned}, navigation~\cite{Chao2014SurveyOpticalFlow}, compression~\cite{Tang2024OfflineOnlineOptical}, and action recognition~\cite{SevillaLara2018integrationopticalflow}.
Due to the advance of technology, the quality and size of videos have been quickly increasing, with current video standards reaching 8K $(7680 \times 4320)$ resolutions.
Nonetheless, most optical flow approaches are designed for only up to 1K resolution inputs and cannot generalize well to larger inputs (\cref{fig:teaser}).
Many methods~\cite{Jiang2021LearningEstimateHidden,Xu2022GMFlowLearningOptical,Zhao2022GlobalMatchingOverlapping} rely on global matching, but their quadratic computational costs limit them.
Transformer-based~\cite{Vaswani2017AttentionIsAll} approaches~\cite{Shi2023FlowFormerMaskedCost,Dong2023RethinkingOpticalFlow,Revaud2023CroCov2Improved} also struggle to adapt to changes in input size~\cite{Huang2022FlowFormerTransformerArchitecture}.
They usually adopt input tiling~\cite{Jaegle2022PerceiverIOgeneral,Huang2022FlowFormerTransformerArchitecture} to overcome the input size limitations, but this strategy creates artifacts due to the loss of global context (\cref{fig:teaser_qualitative}).
Some other methods are not resource-restricted and can theoretically process inputs of larger sizes~\cite{Teed2020RAFTrecurrentall,Xu2021HighResolutionOptical,Wang2024SEARAFTSimple}.
Nonetheless, they show significant estimation degradation with high-resolution inputs.
This problem can be mitigated with high-resolution training~\cite{Xu2021HighResolutionOptical}.
However, this strategy is impractical for very large inputs.

Some existing methods have been shown to work with inputs of up to 4K resolution~\cite{Xu2021HighResolutionOptical,Zheng2022DIPDeepInverse}.
However, they are only qualitatively evaluated on a few hand-picked samples, which does not offer enough evidence of their real generalization capabilities to changes in input size.
Existing optical flow benchmarks only offer up to 2K densely annotated inputs~\cite{Kondermann2016HCIBenchmarkSuite,Mehl2023SpringHighResolution,Shugrina2019CreativeFlow+Dataset}.
TAP-Vid~\cite{Doersch2022TAPVidBenchmark} has some 4K resolution samples, but they only provide annotations for a small set of sparse points per image.
Therefore, there is a lack of quantitative benchmarks to evaluate the performance of optical flow methods on high-resolution inputs.

This paper fills these gaps in high-resolution optical flow estimation in two ways.
First, we propose DPFlow, an optical flow architecture that uses a dual-pyramid bidirectional recurrent encoder to adapt to inputs of multiple sizes.
Our method can be trained only with low-resolution samples and generalize well to 8K resolution inputs.
DPFlow achieves state-of-the-art results on multiple public benchmarks.
We achieve the best zero-shot and fine-tuning evaluation results on the higher-resolution Spring benchmark~\cite{Mehl2023SpringHighResolution}, outperforming SEA-RAFT~\cite{Wang2024SEARAFTSimple} and RPKNet~\cite{Morimitsu2024RecurrentPartialKernel} by $6.5\%$ and $10.8\%$, respectively.
DPFlow also performs well on standard-resolution datasets, achieving the best overall ranking in the MPI-Sintel~\cite{Butler2012naturalisticopensource} and the KITTI 2015~\cite{Menze2015Objectsceneflow} benchmarks.

Second, we created the Kubric-NK dataset, featuring 2400 densely annotated samples at four resolutions ranging from 1K to 8K.
To the best of our knowledge, this is the first dataset to provide a platform for systematically evaluating the performance of optical flow methods with up to extremely large inputs.
We use Kubric-NK to evaluate multiple existing optical flow methods and reveal new findings about how high-resolution inputs affect the predictions.
In particular, we show that DPFlow has strong generalization capabilities, outperforming the next best method by more than $30\%$ at 8K resolution (\cref{fig:teaser_plot}).

In summary, our contributions are as follows:

\begin{enumerate}
    \item We introduce DPFlow, an adaptive optical flow method that only uses standard-resolution samples for training but generalizes well to a wide range of input sizes.
    \item We propose a recurrent bidirectional dual-pyramid encoder that combines variable image- and feature-pyramids to extract multi-scale features from inputs of different sizes.
    \item We create Kubric-NK, a dataset containing multi-resolution samples to push the boundaries of existing optical flow methods by evaluating them with inputs up to 8K resolution.
    \item We demonstrate in extensive experiments that our DPFlow model achieves state-of-the-art optical flow prediction on multiple benchmarks, especially with higher-resolution inputs.
\end{enumerate}

\section{Related Works}
\label{sec:related}

\subsection{Optical Flow Estimation}

The use of multi-scale features for estimating optical flow in a coarse-to-fine manner has been a popular strategy adopted by classical approaches~\cite{Brox2004Highaccuracyoptical,Brox2011LargeDisplacementOptical} to handle large motions.
This strategy has been adopted by multiple models to solve complex problems~\cite{Dosovitskiy2015FlowNetlearningoptical,Ranjan2017Opticalflowestimation,Hui2018LiteFlowNetlightweightconvolutional,Sun2018PWCNetCNNs,Zhang2023aArbitraryshapetext,Sun2022Coarsefinefeature}.
Later, RAFT~\cite{Teed2020RAFTrecurrentall} introduced a paradigm shift by pairing single-scale features with iterative decoding to produce more accurate estimations.
Since then, this architecture has become the foundation of most current approaches.
Some methods adopted global matching strategies~\cite{Xu2022GMFlowLearningOptical,Zhao2022GlobalMatchingOverlapping,Xu2023UnifyingFlowStereo} to produce more reliable initial estimations at fast speeds.
The use of the attention mechanism~\cite{Vaswani2017AttentionIsAll} has also been adopted by multiple models~\cite{Huang2022FlowFormerTransformerArchitecture,Jiang2021LearningEstimateHidden,Revaud2023CroCov2Improved,Dong2023RethinkingOpticalFlow,Kumar2024Evolutiontransformerbased,Xu2023UnifyingFlowStereo,Zhao2022GlobalMatchingOverlapping,Luo2023GAFlowIncorporatingGaussian} to encode more discriminative features.
Some recent works have focused on exploring multi-frame strategies to handle more challenging scenes~\cite{Shi2023VideoFlowExploitingTemporal,Dong2024MemFlowOpticalFlow,Lu2023TransFlowTransformeras}, significantly improving the estimation quality on longer sequences.
However, these methods usually do not perform well when handling inputs whose sizes differ significantly from the training samples.

\subsection{High-Resolution Approaches}

Acknowledging the increasing demand for high-quality videos, a few optical flow methods have proposed ways to handle high-resolution inputs.
Splitting the image with input tiling~\cite{Jaegle2022PerceiverIOgeneral,Huang2022FlowFormerTransformerArchitecture} has been adopted by some top-performing methods~\cite{Shi2023FlowFormerMaskedCost,Dong2023RethinkingOpticalFlow,Revaud2023CroCov2Improved} to get around their size limitations.
Nonetheless, this strategy significantly increases latency and loses global contextual information.
Flow1D~\cite{Xu2021HighResolutionOptical} used 1D decomposition to avoid the quadratic cost of the 4D correlation stage to enable high-resolution processing.
DIP~\cite{Zheng2022DIPDeepInverse} adopted local PatchMatch~\cite{Barnes2009PatchMatchrandomizedcorrespondence} with multi-scale features to replace the global matching stage and reduce computational costs.
Recurrent networks have shown to be a flexible way to adapt to changes in input size~\cite{Sim2021XVFIeXtremeVideo,Morimitsu2024RAPIDFlowRecurrentAdaptable}, although they have not been extensively evaluated for large optical flow prediction tasks.
Based on these previous findings, DPFlow also adopts a recurrent architecture and further improves it with our dual-pyramid approach.

\subsection{Optical Flow Datasets}

Most optical flow datasets are composed of synthetic samples whose annotations can be automatically collected from the rendering engines.
However, even synthetic optical flow datasets usually only provide low-resolution samples compared to current high-definition video standards.
The samples from most commonly used optical flow datasets~\cite{Butler2012naturalisticopensource,Menze2015Objectsceneflow,Wang2020TartanAirDatasetPush,Dosovitskiy2015FlowNetlearningoptical,Mayer2016LargeDatasetTrain} are not larger than around 1000 (1K) pixels wide.
Acknowledging the lack of higher-resolution data, some other datasets~\cite{Mehl2023SpringHighResolution,Kondermann2016HCIBenchmarkSuite,Richter2017PlayingBenchmarks,Shugrina2019CreativeFlow+Dataset,Janai2017SlowFlowExploiting} provided samples with around double the resolution.
While datasets with even larger inputs are available, they have a very limited amount of annotated data.
For example, Middlebury-ST~\cite{Scharstein2014Highresolutionstereo} provided only thirty-three samples at 3K resolution for stereo matching.
TAP-Vid~\cite{Doersch2022TAPVidBenchmark} provides some 4K resolution samples, but they only annotated a very sparse set of around twenty points per video.
Therefore, there is a lack of data for evaluating the performance of optical flow models at high resolutions.
\section{DPFlow}
\label{sec:method}

\begin{figure}[t]
	\centering
	\includegraphics[width=\linewidth]{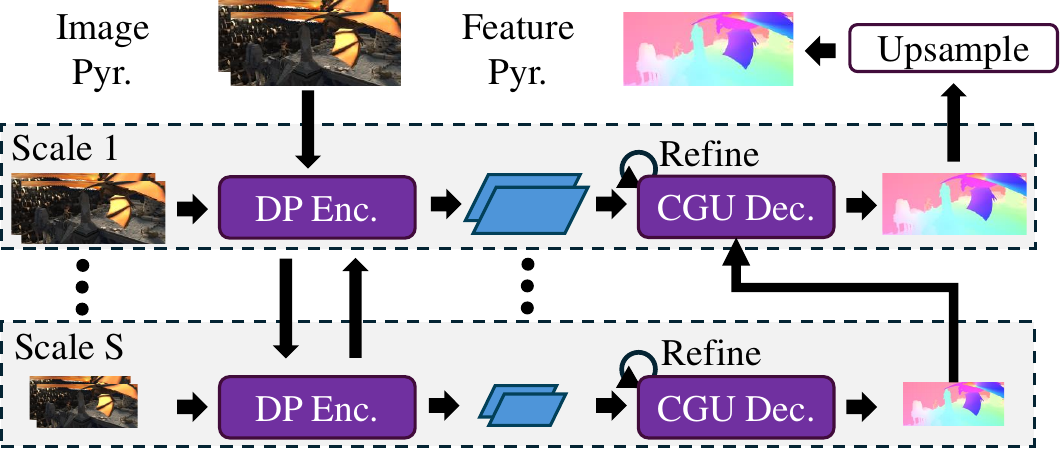}
	\caption{Overview of DPFlow architecture. The dual-pyramid encoder (DP Enc.) combines image and feature pyramids to extract multi-scale features, which the CGU Decoder iteratively refines to produce accurate optical flow predictions.}
	\label{fig:dpflow}
\end{figure}

We adopt a recurrent encoder-decoder structure for optical flow prediction, similar to RAPIDFlow~\cite{Morimitsu2024RAPIDFlowRecurrentAdaptable} and RPKNet~\cite{Morimitsu2024RecurrentPartialKernel}.
However, we propose a novel recurrent dual-pyramid encoder to extract multi-scale features.
This encoder builds on RPKNet and combines the advantages of image and feature pyramids in a flexible bidirectional recurrent structure.
DPFlow employs a convolutional Cross-Gated Unit (CGU) to modulate the information flow to collect discriminative features at lower costs.
Our decoder follows the structure of RAPIDFlow, but we adopt a GRU equipped with our CGU as the main refinement block.
DPFlow's structure is shown in~\cref{fig:dpflow}.

\subsection{Dual-Pyramid Encoder}

\begin{figure}[t]
    \centering
    \begin{subfigure}{0.35\linewidth}
        \centering
        \includegraphics[width=\linewidth]{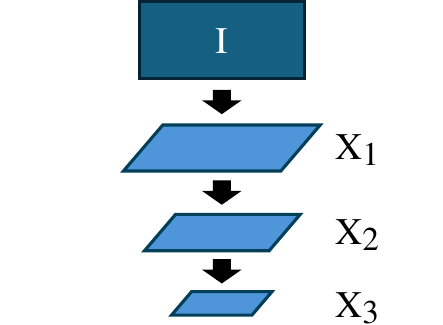}
        \caption{$X_s = \phi^s(I)$}
    \end{subfigure}
    \qquad
    \begin{subfigure}{0.35\linewidth}
        \centering
        \includegraphics[width=\linewidth]{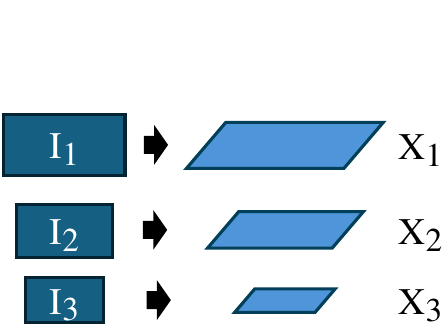}
        \caption{$X_s = \phi(I_s)$}
    \end{subfigure}
    \caption{(a) Feature pyramid. (b) Image pyramid. $\phi$ represents a network block, $I$ denotes images and $X$ features.}
    \label{fig:pyramids}
\end{figure}

Most methods use image or feature pyramids to extract multi-scale features from the inputs (\cref{fig:pyramids}).
Image pyramids produce a single feature per level and provide a more direct path to the input information.
On the other hand, feature pyramids transmit multi-scale information across levels, but the input information may be diluted in the deepest levels.
Pyramids become even more challenging with a recurrent network that shares the same block across all levels, which may cause the gradients to vanish during training.

\begin{figure}[t]
	\centering
	\includegraphics[width=\linewidth]{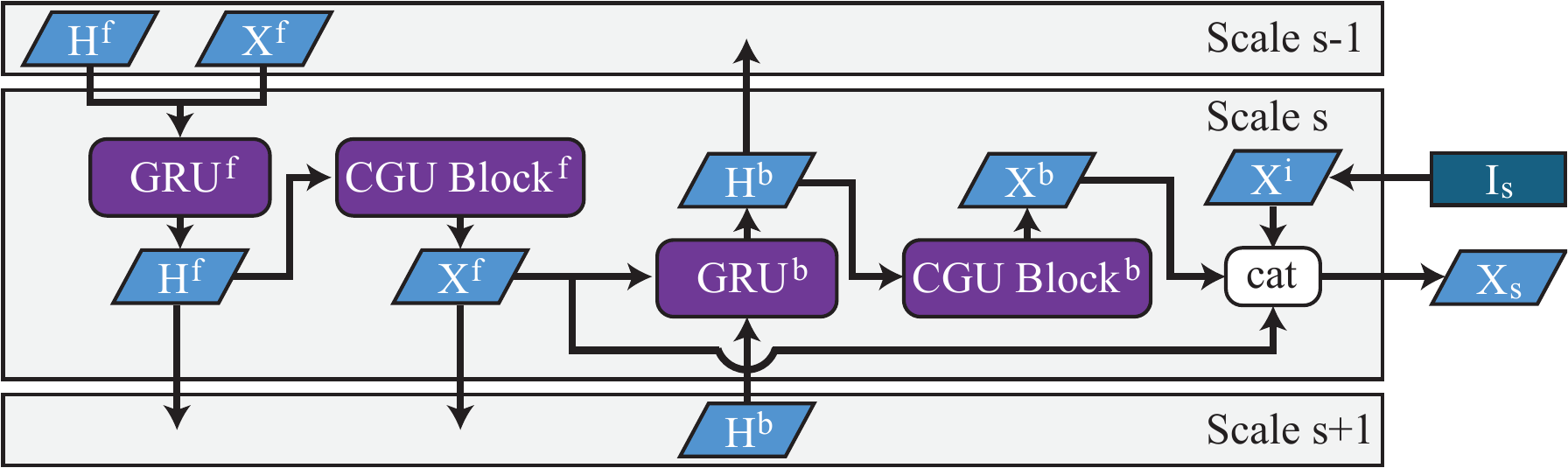}
	\caption{Illustration of the dual-pyramid encoder block.}
	\label{fig:encoder_block}
\end{figure}

We propose to combine the properties of both pyramid approaches in a unified dual-pyramid encoder.
This allows us to share multi-scale information across levels while using the image pyramid to retain access to input information at all levels.
Our recurrent network adopts ConvGRU~\cite{Ballas2016DelvingDeeperConvolutional} to avoid training instability and better modulate the information flow across levels.
Moreover, we adopt a bidirectional recurrent unit to allow shallower levels access to information from deeper levels.
The pipeline of our recurrent dual-pyramid network is illustrated in~\cref{fig:encoder_block} and described as:
    
\begin{equation}
	\begin{aligned}
		X_0^f, H_0^f &= \phi^{\text{stem}}(I),\\
		H_s^f &= \text{ConvGRU}^f(H_{s-1}^f, X_{s-1}^f),\\
		X_s^f &= \text{CGU Block}^f(H_s^f),\\
		H_s^b &= \text{ConvGRU}^b(H_{s+1}^b, X_s^f),\\
		X_s^b &= \text{CGU Block}^b(H_s^b),\\
		X_s^i &= \phi^i(I_s),\\
        X_s &= \phi^{\text{out}}(\text{concat}(X_s^f, X_s^b, X_s^i)),
	\end{aligned}
    \label{eq:dual_pyramid}
\end{equation}
where CGU Block is our Cross-Gated Unit block, $I$ represents the original input image, and $I_s$ is downsampled to the scale $s$.
$\phi$ denotes a convolution block and the superscripts $f, b, i$ denote forward, backward, and image, respectively.
We initialize the state $H^b$ as zero, and the output feature $X_s$ is obtained by concatenating the intermediate forward, backward, and image pyramid features.
The same process is applied to both images of the input pair.

\subsection{Cross-Gated Unit (CGU)}

\begin{figure}[t]
	\centering
	\includegraphics[width=\linewidth]{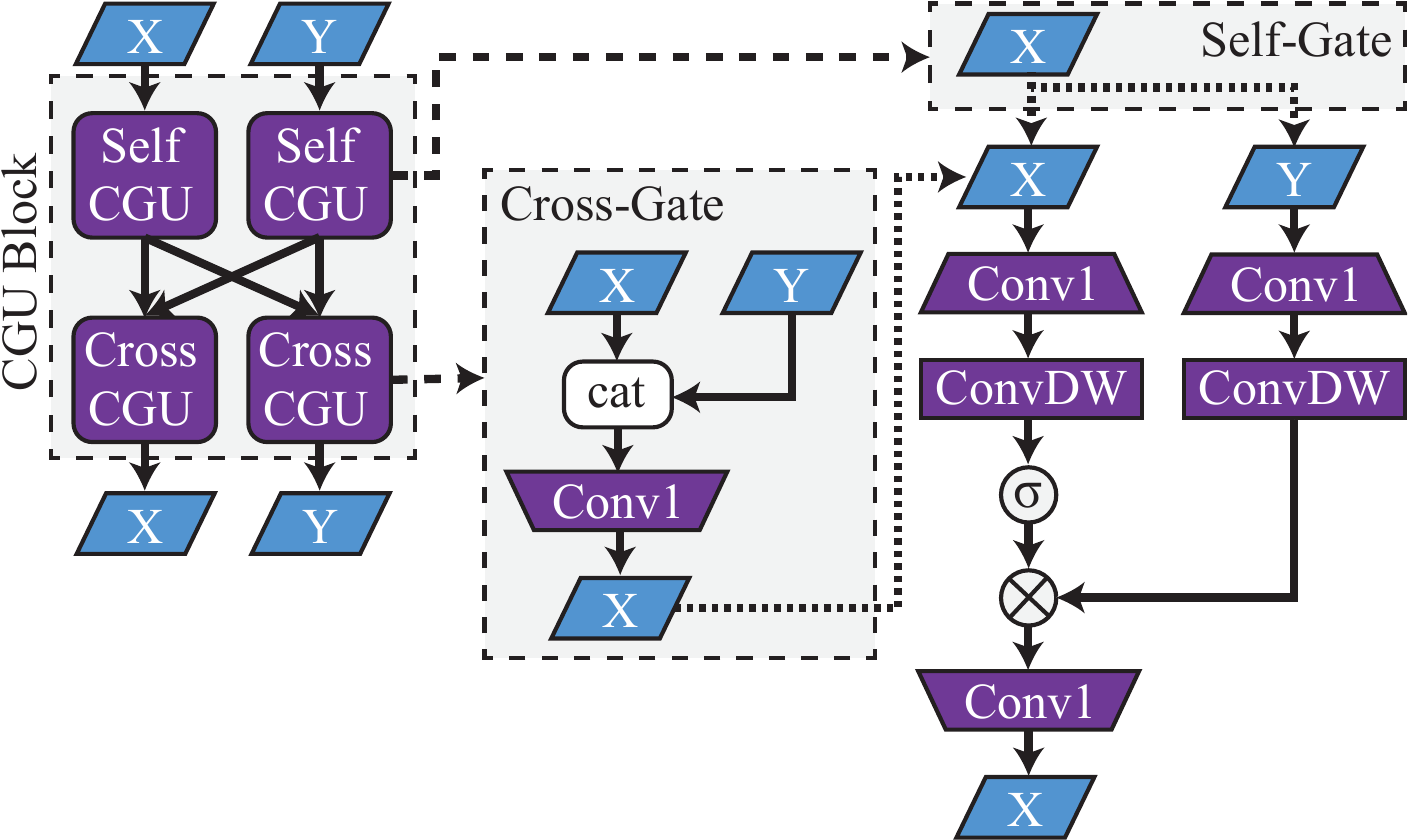}
	\caption{Illustration of the CGU block, composed of convolutional self- and cross-gates. Conv1 uses a $1 \times 1$ kernel, and ConvDW denotes depthwise convolution. We adopt GELU~\cite{Hendrycks2016GaussianErrorLinear} as the non-linear gate $\sigma$ and Layer Scale~\cite{Touvron2021Goingdeeperimage} to modulate the residual.}
	\label{fig:cgu}
\end{figure}

Most models adopt self- and cross-attention~\cite{Vaswani2017AttentionIsAll,Xu2022GMFlowLearningOptical,Sun2021LoFTRDetectorFree} to extract more discriminative features for matching.
However, attention mechanisms pose challenges, such as high computational costs and lack of generalization to different input sizes~\cite{Jaegle2022PerceiverIOgeneral,Huang2022FlowFormerTransformerArchitecture}.
CNNs avoid these problems using simpler local operations without requiring explicit positional encoding~\cite{Islam2024Positionpaddingpredictions}.
Therefore, our DPFlow model adopts a pure convolution-based Cross-Gated Unit (CGU) as its main component.
The CGU structure is illustrated in~\cref{fig:cgu}.
Its design is inspired by the Gated-CNN block~\cite{Dauphin2017LanguageModelingGated} with a cross-gate to leverage information from a pair of inputs.
If only one input is provided, the CGU acts as a self-gate.

\subsection{Loss Function}

We train our model using the Mixture-of-Laplace loss~\cite{Wang2024SEARAFTSimple} $(\mathcal{L}_{\text{MoL}})$ but adapted for multi-scale training.
The MoL loss reduces the impact of points with ambiguous predictions that can otherwise dominate the loss term.
We train our models using three scales and four iterations per scale.
However, these settings can be freely changed during inference.
Our loss function is hence defined as:
\begin{equation}
    \mathcal{L}(F_{s, k}^{\text{pred}}, F) = \sum_{s=1}^{3}{\sum_{k=1}^{4}{\gamma^{4s-k} \mathcal{L}_{\text{MoL}}(F_{s, k}^{\text{pred}}, F)}},
\end{equation}
where $\gamma$ is a scalar weight, $F$ is the groundtruth flow and $F_{s, k}^{\text{pred}}$ is the prediction at scale $s$ and refinement iteration $k$.
\section{Kubric-NK Dataset}
\label{sec:dataset}

\begin{figure}[t]
    \centering
    \begin{subfigure}{\linewidth}
        \centering
        \includegraphics[width=\linewidth]{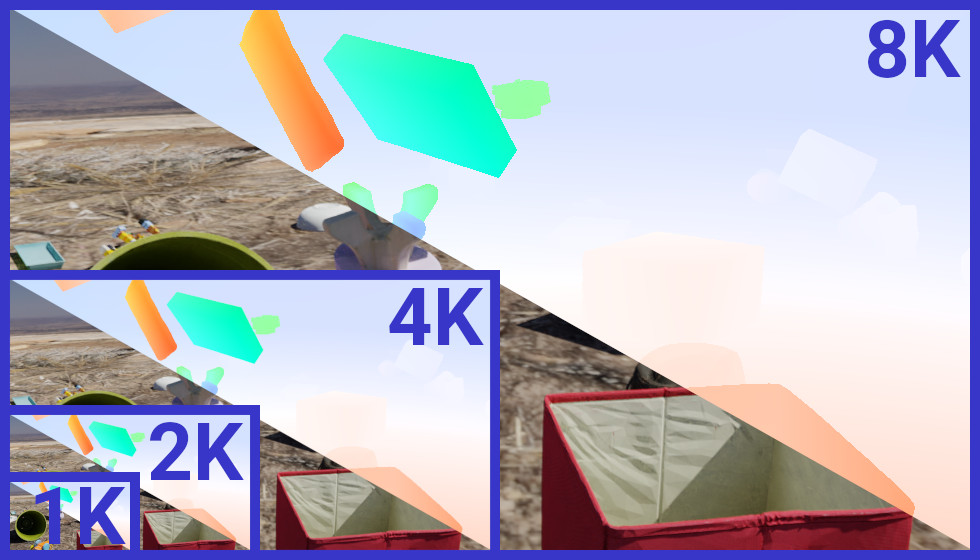}
        \caption{Image pair and optical flow}
        \label{fig:kubric_flow}
    \end{subfigure}
    \begin{subfigure}{0.32\linewidth}
        \centering
        \includegraphics[width=\linewidth]{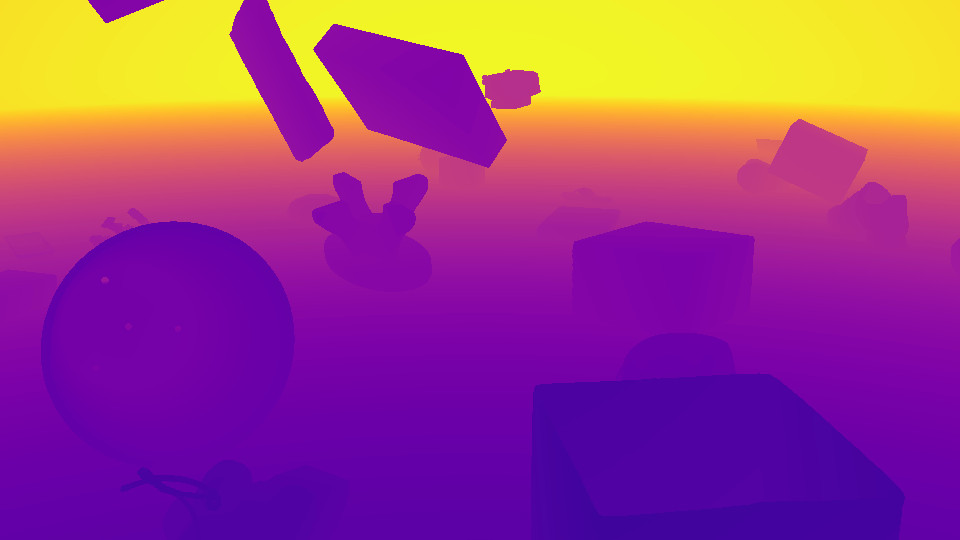}
        \caption{Depth}
        \label{fig:kubric_depth}
    \end{subfigure}
    \begin{subfigure}{0.32\linewidth}
        \centering
        \includegraphics[width=\linewidth]{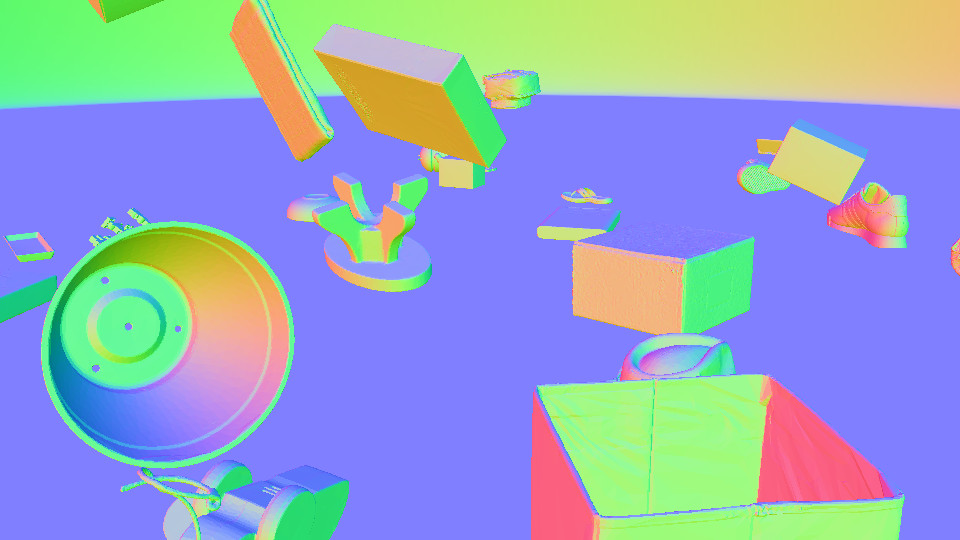}
        \caption{Surface normals}
        \label{fig:kubric_normal}
    \end{subfigure}
    \begin{subfigure}{0.32\linewidth}
        \centering
        \includegraphics[width=\linewidth]{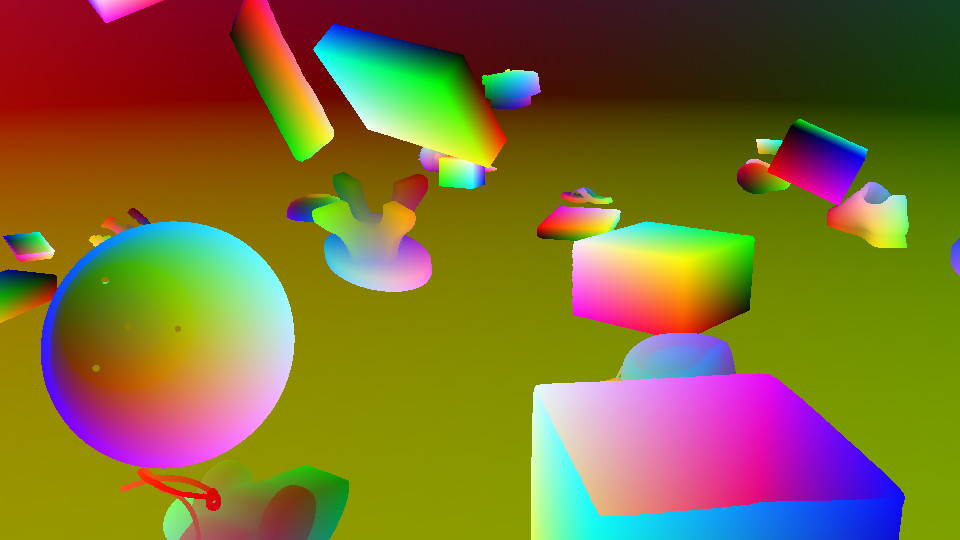}
        \caption{Object coordinates}
        \label{fig:kubric_object_coordinates}
    \end{subfigure}
    \caption{The Kubric-NK dataset provides the same samples from 1K to 8K resolutions to test generalization to different input sizes.}
    \label{fig:kubric_nk}
\end{figure}

\begin{figure}[t]
    \centering
    \includegraphics[width=\linewidth]{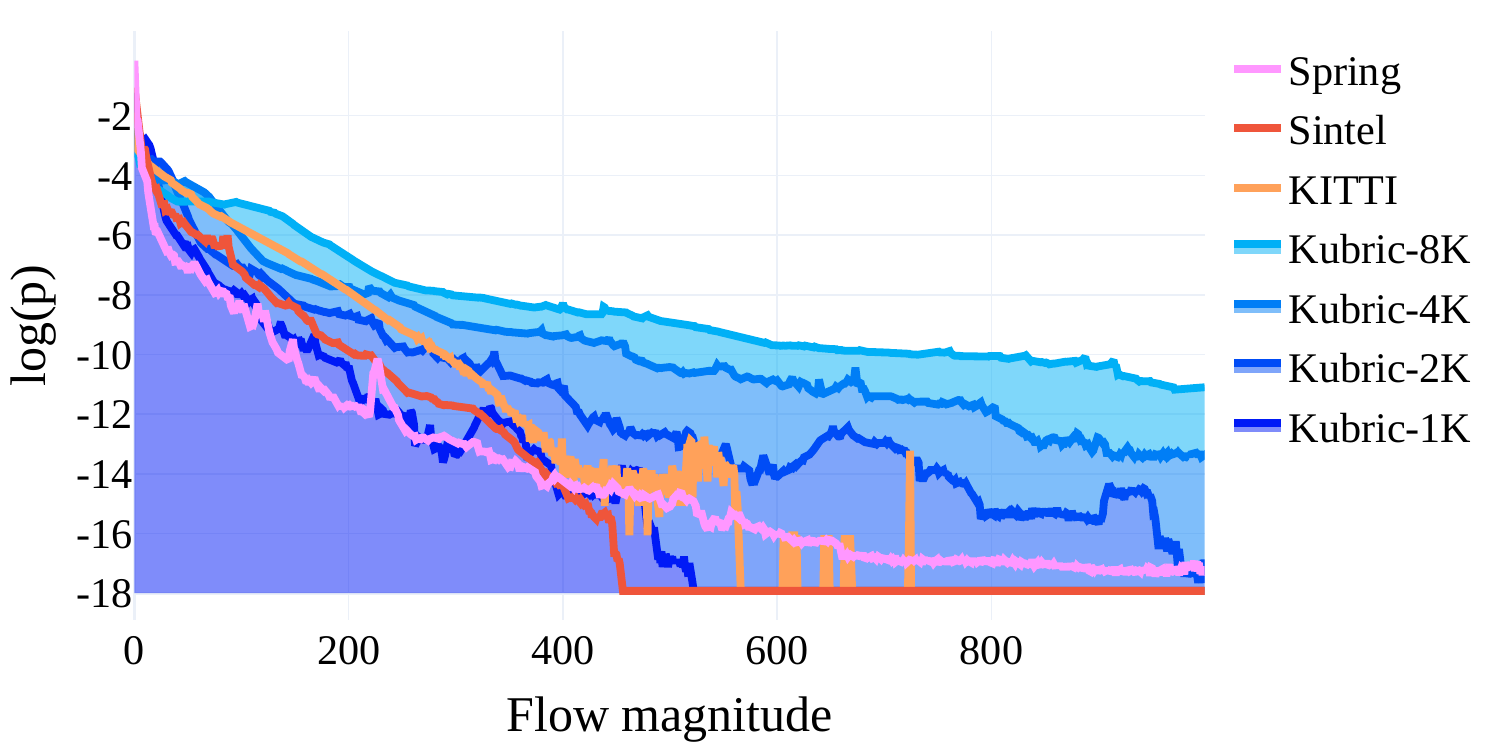}
    \caption{Comparison of flow magnitude distributions between Kubric-NK and other datasets.}
    \label{fig:flow_magnitudes}
\end{figure}

Optical flow methods require resolution generalization as flow vectors correlate with input spatial resolution.
Nonetheless, current optical flow evaluation datasets do not provide a way to evaluate the robustness of optical flow methods to changes in resolution.
Most papers~\cite{Zheng2022DIPDeepInverse,Xu2021HighResolutionOptical,Seo2024LightweightOpticalFlow} show generalization to high resolutions with only a few qualitative examples, which makes it hard to compare different methods.
In this paper, we employ Kubric~\cite{Greff2022Kubricscalabledataset} to generate a multi-resolution synthetic optical flow dataset.
We procedurally generate 30 sequences containing 600 unique densely annotated samples at four resolutions: 1K, 2K, 4K, and 8K.
In addition to the optical flow, other dense annotations, such as depth, surface normals, and object coordinates, are also available.
We define the input resolution $n$K as $(960n, 540n)$ and thus refer to our dataset as Kubric-NK.
\Cref{fig:kubric_nk} shows some samples from the proposed Kubric-NK dataset.

This dataset serves two main purposes: (1) to provide a common platform to quantitatively evaluate the performance of optical flow methods at high resolutions and (2) to analyze how input size changes affect different models by providing the same scenes rendered at multiple resolutions.
The histograms in~\cref{fig:flow_magnitudes} show that Kubric-NK optical flow samples at 1K resolution have a similar distribution to other popular datasets, such as MPI-Sintel~\cite{Butler2012naturalisticopensource}.
At the same time, Kubric-NK's provides significantly larger and more challenging samples at higher resolutions.
\section{Experimental Results}

\subsection{Dataset and Implementation}
\label{sec:implementation}

We follow the traditional optical flow training pipeline of five stages.
We do not use any additional high-resolution datasets for training.
Stage 1 comprises 100k iterations on the FlyingChairs dataset~\cite{Dosovitskiy2015FlowNetlearningoptical}.
Stage 2 adds 1M more iterations on the FlyingThings3D~\cite{Mayer2016LargeDatasetTrain}.
For stage 3, we use 120k iterations to finetune the model in a mixed dataset combining FlyingThings3D, KITTI 2015~\cite{Menze2015Objectsceneflow}, HD1K~\cite{Kondermann2016HCIBenchmarkSuite}, and Sintel~\cite{Butler2012naturalisticopensource} samples.
Stage 4a further finetunes the model in the KITTI 2015 dataset for 2k iterations.
Stage 4b finetunes the model from stage 3 for 120k more iterations using the Spring~\cite{Mehl2023SpringHighResolution} training samples.
Our models are implemented in PyTorch and trained on two NVIDIA RTX3090 GPUs.

We evaluate the results on each dataset using their official metrics.
MPI-Sintel, Middlebury-ST~\cite{Scharstein2014Highresolutionstereo}, and Kubric-NK adopt the End-Point-Error (EPE).
Spring is evaluated with the 1-pixel (1px) metric, KITTI 2015 uses the Fl-All metric, while VIPER~\cite{Richter2017PlayingBenchmarks} uses their proposed WAUC metric.
Additional details are provided in the Sup. Mat.

\subsection{High-Resolution Evaluation}

Previous works usually only provide qualitative evaluation for generalization to high-resolution inputs~\cite{Zheng2022DIPDeepInverse,Xu2021HighResolutionOptical}.
Here, we provide quantitative metrics of current optical flow models on various existing high-resolution samples and our Kubric-NK dataset.
\Cref{tab:high_res_datasets} provides an overview of the datasets we use in our evaluation.
The results are collected using open-source implementations of the models~\cite{Revaud2023CroCov2Improved,Morimitsu2021PyTorchLightningOptical}.
DPFlow automatically chooses the number of pyramid levels $N$ based on the diagonal input size: $N = \text{round}(\log_2(\max(1, \sqrt{W^2 + H^2} / \sqrt{960^2 + 540^2}))) + 3$ and adopts 4 refinement iterations per level.

\subsubsection{Evaluation Checkpoint}

Before testing each model, we must choose which training stage to use for the experiments.
We evaluate the results of models trained with stages 2 and 3 on a subset of our chosen high-resolution datasets to check how each checkpoint behaves.
\Cref{tab:ckpt_comparison} shows the average error obtained by a set of methods trained with each stage.
Due to their large sizes, we use only $10\%$ of Spring and VIPER samples for this evaluation.
The selected methods list and their results are provided in the Sup. Mat.
The results indicate that stage 3 training is the best choice, except for Middlebury-ST.
Stage 3's underperformance may be due to the low number of samples and the stereo images.
We use the results from~\cref{tab:ckpt_comparison} to choose the training checkpoint for each benchmark.

\begin{table}[t]
    \caption{Overview of the datasets used in our high-resolution experiments.}
    \centering
    \begin{tabular}{@{}lcccc@{}}
        \toprule
        Dataset & Resolution & Train & Test\\
        \midrule
        Spring~\cite{Mehl2023SpringHighResolution} & 2K (4K GT) & 4963 & 3960\\
        Midbury-ST~\cite{Scharstein2014Highresolutionstereo} & 3K & 23 & -\\
        VIPER~\cite{Richter2017PlayingBenchmarks} & 2K & 4954 & -\\
        Kubric-NK & 1K, 2K, 4K, 8K & - & $4 \times 600$\\
        \bottomrule
    \end{tabular}
    \label{tab:high_res_datasets}
\end{table}

\begin{table}[t]
    \caption{Average results with different training stages. Results in bold indicate which stage we used for each dataset. }
    \centering
    \begin{tabular}{@{}llcc@{}}
        \toprule
        Dataset & Metric & Stage 2 & Stage 3\\
        \midrule
        Spring~\cite{Mehl2023SpringHighResolution} & 1px $\downarrow$ & 4.92 & \textbf{4.53}\\
        VIPER~\cite{Richter2017PlayingBenchmarks} & WAUC $\uparrow$ & 63.1 & \textbf{65.9}\\
        Middlebury-ST~\cite{Scharstein2014Highresolutionstereo} & EPE $\downarrow$ & \textbf{24.3} & 32.9\\
        Kubric-1K & EPE $\downarrow$ & 0.72 & \textbf{0.70}\\
        \bottomrule
    \end{tabular}
    \label{tab:ckpt_comparison}
\end{table}

\begin{table*}[t]
	\caption{Results on high-resolution datasets. In Spring, ft. indicates results obtained after fine-tuning the model with Spring samples, while 0-sh. is the zero-shot evaluation. $\dag$ results are not available due to running out of memory. $^{\Box}$ denotes methods that use input tiling~\cite{Jaegle2022PerceiverIOgeneral,Huang2022FlowFormerTransformerArchitecture} and $^\ast$ methods with adaptive pyramids. The metrics of each dataset are explained in~\cref{sec:implementation} and in the Sup. Mat.}
	\centering
	\begin{tabular}{@{}llccccccccccc@{}}
		\toprule
		   &  & \multicolumn{4}{c}{Spring (1px $\downarrow$)} & VIPER & Mbury-ST & \multicolumn{4}{c}{Kubric-NK}\\
		\cmidrule(lr){3-6}
		  Method & Venue & \multicolumn{2}{c}{train} & \multicolumn{2}{c}{test-2K} & WAUC $\uparrow$ & EPE $\downarrow$ & \multicolumn{4}{c}{EPE $\downarrow$}\\
            \cmidrule(lr){3-4} \cmidrule(lr){5-6} \cmidrule(lr){9-12}
             &  & 2K & 4K & ft. & 0-sh. & 2K & 3K & 1K & 2K & 4K & 8K\\
		\midrule
            PWC-Net~\cite{Sun2018PWCNetCNNs} & CVPR'18 & 4.94 & 9.78 & - & 82.2 & 66.0 & 35.9 & 1.26 & 3.11 & 11.4& 40.8\\
            IRR~\cite{Hur2019Iterativeresidualrefinement} & CVPR'19 & 3.85 & 8.53 & - & - & 66.3 & 15.2 & 1.09 & 2.26 & 15.6 & $\dag$\\
            VCN~\cite{Yang2019Volumetriccorrespondencenetworks} & NeurIPS'19 & 7.50 & - & - & - & 62.9 & 44.2 & 1.08 & 2.77 & $\dag$ & $\dag$\\
            RAFT~\cite{Teed2020RAFTrecurrentall} & ECCV'20 & 3.85 & 10.7 & - & 6.79 & 71.1 & 35.2 & 0.68 & 1.46 & 12.3 & 82.7\\
            GMA~\cite{Jiang2021LearningEstimateHidden} & ICCV'21 & 3.75 & $\dag$ & - & 7.07 & 71.7 & $\dag$ & 0.66 & 1.30 & $\dag$ & $\dag$\\ 
            DIP~\cite{Zheng2022DIPDeepInverse} & CVPR'22 & 3.64 & 9.53 & - & - & 71.3 & 16.7 & 0.68 & 1.57 & 5.12 & $\dag$\\
            GMFlow~\cite{Xu2022GMFlowLearningOptical} & CVPR'22 & 6.71 & $\dag$ & - & 10.35 & 56.7 & $\dag$ & 0.88 & 1.38 & $\dag$ & $\dag$\\ 
            FlowFormer~\cite{Huang2022FlowFormerTransformerArchitecture}$^{\Box}$ & ECCV'22 & 3.78 & 9.53 & - & 6.51 & 72.7 & 19.0 & 0.52 & 1.04 & 5.60 & 24.4\\
            SKFlow~\cite{Sun2022SKFlowLearningOptical} & NeurIPS'22 & 3.79 & $\dag$ & - & - & 71.9 & $\dag$ & 0.62 & 1.07 & $\dag$ & $\dag$\\ 
            FlowFormer++~\cite{Shi2023FlowFormerMaskedCost}$^{\Box}$ & CVPR'23 & 3.76 & 9.71 & - & - & 72.5 & 17.5 & 0.46 & 1.11 & 5.39 & 24.4\\ 
            MatchFlow~\cite{Dong2023RethinkingOpticalFlow}$^{\Box}$ & CVPR'23 & 3.97 & 9.32 & - & - & 73.0 & 16.7 & 0.54 & 1.25 & 6.93 & 29.0\\ 
            CroCo-Flow~\cite{Revaud2023CroCov2Improved}$^{\Box}$ & ICCV'23 & 3.51 & $\dag$ & 4.56 & - & \textbf{76.1} & $\dag$ & \textbf{0.45} & 1.06 & $\dag$ & $\dag$\\ 
            MS-RAFT+~\cite{Jahedi2023MSRAFTHigh} & IJCV'23 & \underline{3.08} & $\dag$ & - & 5.72 & - & $\dag$ & 0.55 & 1.27 & $\dag$ & $\dag$\\ 
            GMFlow+~\cite{Xu2023UnifyingFlowStereo} & TPAMI'23 & 5.72 & $\dag$ & - & - & 69.3 & $\dag$ & \textbf{0.45} & \underline{0.97} & $\dag$ & $\dag$\\ 
            RPKNet~\cite{Morimitsu2024RecurrentPartialKernel} & AAAI'24 & 3.28 & 7.35 & - & \underline{4.80} & 72.7 & 15.0 & 0.60 & 1.05 & 3.96 & 18.2\\ 
            MemFlow~\cite{Dong2024MemFlowOpticalFlow} & CVPR'24 & 3.24 & $\dag$ & 4.48 & 5.76 & 74.6 & $\dag$ & 0.55 & \underline{0.97} & $\dag$ & $\dag$\\
            SEA-RAFT (M)~\cite{Wang2024SEARAFTSimple} & ECCV'24 & 3.47 & 8.22 & \underline{3.68} & - & \underline{75.5} & 65.1 & 0.53 & 1.53 & 11.0 & 51.8\\
            RAPIDFlow~\cite{Morimitsu2024RAPIDFlowRecurrentAdaptable}$^\ast$ & ICRA'24 & 3.73 & \underline{6.91} & - & - & 71.0 & \underline{6.41} & 0.81 & 1.23 & \underline{2.11} & \underline{5.96}\\
            CCMR+~\cite{Jahedi2024CCMRHighResolution} & WACV'24 & 3.27 & $\dag$ & - & - & 75.2 & $\dag$ & 0.54 & 1.41 & $\dag$ & $\dag$\\ 
            DPFlow$^\ast$ &  & \textbf{3.06} & \textbf{6.03} & \textbf{3.44} & \textbf{4.28} & 72.9 & \textbf{5.03} & 0.50 & \textbf{0.85} & \textbf{1.77} & \textbf{4.11}\\
		\bottomrule
        \end{tabular}
	\label{tab:sota_hd}
\end{table*}

\subsubsection{Spring Dataset}

We conduct three tests with the Spring benchmark.
The first test evaluates the model generalization using the train split of the dataset.
Since Spring provides 4K annotations, our second test uses bicubic interpolation to double the resolution of the images and repeat the previous experiment (Spring 4K).
For the last test, we used training stages 3 and 4b and submitted our results to the public zero-shot and finetuning benchmarks, respectively.
\Cref{tab:sota_hd} shows that DPFlow achieves the best results in all variants of the Spring benchmark.
Regarding the public test results, DPFlow outperforms previous methods by over $6.5\%$ in the zero-shot and fine-tuning settings.
The advantage of our method becomes even more apparent in the 4K evaluation, where we outperform RAPIDFlow by $13\%$, thus demonstrating the advantage of our more robust dual-pyramid encoder.
Compared to fixed-pyramid methods like SEA-RAFT~\cite{Wang2024SEARAFTSimple}, we achieve an even more notable improvement of $27\%$.

\subsubsection{VIPER}

The VIPER dataset provides FullHD resolution samples captured by a driving car in a video game.
At this resolution, most methods produce reliable predictions.
DPFlow performs on par with MatchFlow using the default pyramid.
However, as shown in~\cref{tab:pyramid}, better results could be achieved by tuning the pyramid levels.

\subsubsection{Middlebury-ST}

This dataset provides high-resolution real images for the task of stereo prediction.
We use the samples with \textit{imperfect} stereo alignments, making the task similar to optical flow estimation.
With 3K resolution samples, the advantages of DPFlow are more evident.
We obtain a $21\%$ error reduction against RAPIDFlow and $66\%$ against RPKNet.

\subsubsection{Kubric-NK}

We use Kubric-NK to analyze the behavior of optical flow models when the input size changes.
The results in~\cref{tab:sota_hd} and~\cref{fig:teaser_plot} reveal some new and important characteristics of optical flow methods.

\textbf{Scaling behavior.}
The input size directly affects the flow magnitude, causing higher errors at larger resolutions.
Methods based on single-resolution features, such as RAFT~\cite{Teed2020RAFTrecurrentall} variants, show exponential degradation when the resolution increases (e.g., SEA-RAFT's estimation error increases by two orders of magnitude from 1K to 8K).
Using multi-scale features (RPKNet, DIP) provides more robustness to changes in input size, but they still do not provide satisfactory predictions at 8K resolution.
Adaptive pyramids (RAPIDFlow, DPFlow) can counteract the resolution increase and stabilize the predictions, keeping a linear error increase according to the input size.
DPFlow outperforms all other approaches at larger resolutions, surpassing RAPIDFlow by over $30\%$ at 8K.

\textbf{Resource management.}
Many top-performing methods cannot handle large 4K and 8K inputs with consumer hardware equipped with 24GB GPUs.
Therefore, developing efficient methods for high-resolution inputs is necessary.

\textbf{Input tiling.}
Cropping the input mitigates the memory overflow and somewhat stabilizes the predictions.
However, cropping loses global information, leading to obvious square-shaped artifacts (see~\cref{fig:teaser_qualitative}) and significantly longer latencies at high resolutions.

\begin{figure*}[t]
	\centering
	\includegraphics[width=\linewidth]{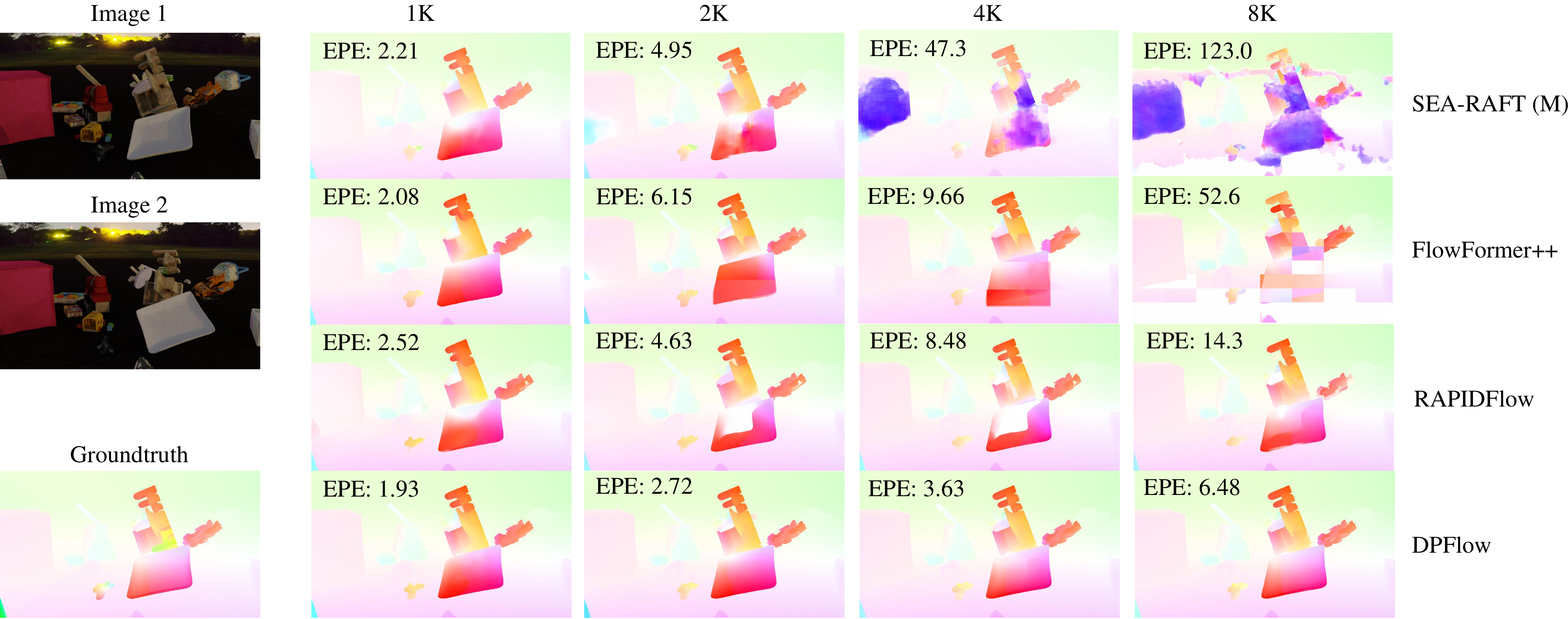}
	\caption{Qualitative results comparing the behavior of some optical flow methods when the input size grows from 1K to 8K resolution.}
	\label{fig:kubric_nk_qualitative}
\end{figure*}

\textbf{Qualitative results.}
\Cref{fig:kubric_nk_qualitative} illustrates how the prediction of some methods changes with increasing resolution inputs.
SEA-RAFT uses fixed-scale features, and it is significantly affected by high-resolution inputs.
The tiling technique adopted by FlowFormer++ mitigates the problem to some extent, but the loss of global information causes noticeable degradation at 8K.
RAPIDFlow shows relatively accurate results due to its adaptive pyramids, but its predictions miss important structural details.
DPFlow maintains more stable results across all resolutions.

\subsection{Standard-Resolution Evaluation}

We also evaluate our model on the widely adopted MPI-Sintel~\cite{Butler2012naturalisticopensource} and KITTI 2015~\cite{Menze2015Objectsceneflow} datasets, which contains samples at around 1K resolution.

\subsubsection{Public Benchmarks}

We use the models trained with stages 3 and 4a to submit our results to the official public benchmarks.
The results in~\cref{tab:sintel_kitti_test} show that DPFlow can also achieve top-performing results in standard resolution.
Among two-frame methods, we achieved the best average rank on the KITTI and Sintel benchmarks.
Also, despite achieving remarkable results with its adaptive pyramid at higher resolutions, we observe that RAPIDFlow~\cite{Morimitsu2024RAPIDFlowRecurrentAdaptable} underperforms at these standard benchmarks.
DPFlow, on the other hand, shows consistent results in all evaluations, highlighting the benefits of our proposed approach.

\subsubsection{Generalization}

We evaluate the generalization capabilities using the model trained until stage two for a zero-shot evaluation on the selected benchmarks.
\Cref{tab:sintel_kitti_train} shows we also achieve remarkable generalization results.
We outperform FlowDiffuser~\cite{Luo2024FlowDiffuserAdvancingOptical} and SEA-RAFT (L)~\cite{Wang2024SEARAFTSimple} by $6\%$ on the KITTI dataset.
DPFlow is also only $3\%$ behind the top-performing FlowDiffuser on the Sintel final pass.

\subsection{Ablation Study}

\begin{table}[t]
	\caption{Benchmark evaluation on MPI-Sintel and KITTI 2015 public test sets. The last column represents the average ranking on the three benchmarks.}
	\centering
	\begin{tabular}{@{}lcccc@{}}
		\toprule
	  \multirow{2}{*}{Method} & \multicolumn{2}{c}{MPI-Sintel} & KITTI 2015 & Avg.\\
		\cmidrule(lr){2-3}
		  & clean & final & Fl-All & rank\\
		\midrule
		RAPIDFlow~\cite{Morimitsu2024RAPIDFlowRecurrentAdaptable} & 2.03 & 3.56 & 6.12 & 21.0\\
		RAFT~\cite{Teed2020RAFTrecurrentall} & 1.60 & 2.85 & 5.10 & 19.3\\
		GMA~\cite{Jiang2021LearningEstimateHidden} & 1.38 & 2.47 & 5.25 & 17.0\\
		RPKNet~\cite{Morimitsu2024RecurrentPartialKernel} & 1.31 & 2.65 & 4.64 & 15.7\\
		DIP~\cite{Zheng2022DIPDeepInverse} & 1.43 & 2.83 & 4.21 & 14.7\\
		KPA-Flow~\cite{Luo2022LearningOpticalFlow} & 1.35 & 2.36 & 4.60 & 13.0\\
		SKFlow~\cite{Sun2022SKFlowLearningOptical} & 1.28 & 2.27 & 4.84 & 13.0\\
		SEA-RAFT (L)~\cite{Wang2024SEARAFTSimple} & 1.31 & 2.60 & 4.30 & 12.7\\
		DDVM~\cite{Saxena2023SurprisingEffectivenessDiffusion} & 1.75 & 2.47 & \textbf{3.26} & 11.7\\
		MS-RAFT+~\cite{Jahedi2023MSRAFTHigh} & 1.23 & 2.68 & 4.19 & 11.7\\
            FlowFormer~\cite{Huang2022FlowFormerTransformerArchitecture} & 1.19 & 2.12 & 4.68 & 11.3\\
		MatchFlow~\cite{Dong2023RethinkingOpticalFlow} & 1.16 & 2.37 & 4.63 & 11.3\\
		AnyFlow~\cite{Jung2023AnyFlowArbitraryScale} & 1.23 & 2.44 & 4.41 & 10.7\\
		GMFlow+~\cite{Xu2023UnifyingFlowStereo} & 1.03 & 2.37 & 4.49 & 8.0\\
		CroCo-Flow~\cite{Revaud2023CroCov2Improved} & 1.09 & 2.44 & 3.64 & 7.3\\
		GAFlow~\cite{Luo2023GAFlowIncorporatingGaussian} & 1.15 & 2.05 & 4.42 & 7.3\\
		FlowFormer++~\cite{Shi2023FlowFormerMaskedCost} & 1.07 & \textbf{1.94} & 4.52 & 6.3\\
		SAMFlow~\cite{Zhou2024SAMFlowEliminatingAny} & \textbf{1.00} & 2.08 & 4.49 & 5.7\\
		CCMR+~\cite{Jahedi2024CCMRHighResolution} & 1.07 & 2.10 & 3.86 & 5.0\\
		FlowDiffuser~\cite{Luo2024FlowDiffuserAdvancingOptical} & \underline{1.02} & 2.03 & 4.17 & \underline{3.3}\\
		DPFlow & 1.04 & \underline{1.97} & \underline{3.56} & \textbf{2.7}\\
		\bottomrule
	\end{tabular}
	\label{tab:sintel_kitti_test}
\end{table}

\begin{table}[t]
	\caption{Generalization results on MPI-Sintel and KITTI 2015 train sets.}
	\centering
	\begin{tabular}{@{}lcccc@{}}
		\toprule
	  \multirow{2}{*}{Method} & \multicolumn{2}{c}{MPI-Sintel} & \multicolumn{2}{c}{KITTI 2015}\\
		\cmidrule(lr){2-3} \cmidrule(lr){4-5}
		  & clean & final & EPE & Fl-All\\
		\midrule
		RAFT~\cite{Teed2020RAFTrecurrentall} & 1.44 & 2.71 & 5.04 & 17.4\\
		GMA~\cite{Jiang2021LearningEstimateHidden} & 1.30 & 2.74 & 4.69 & 17.1\\
		DIP~\cite{Zheng2022DIPDeepInverse} & 1.30 & 2.82 & 4.29 & 13.7\\
		FlowFormer~\cite{Huang2022FlowFormerTransformerArchitecture} & 0.95 & 2.35 & 4.09 & 14.7\\
		GMFlowNet~\cite{Zhao2022GlobalMatchingOverlapping} & 1.14 & 2.71 & 4.24 & 15.4\\
		SKFlow~\cite{Sun2022SKFlowLearningOptical} & 1.22 & 2.46 & 4.47 & 15.5\\
		AnyFlow~\cite{Jung2023AnyFlowArbitraryScale} & 1.10 & 2.52 & 3.76 & 12.4\\
		FlowFormer++~\cite{Shi2023FlowFormerMaskedCost} & \underline{0.90} & 2.30 & 3.93 & 14.1\\
		GAFlow~\cite{Luo2023GAFlowIncorporatingGaussian} & 0.95 & 2.34 & 3.92 & 13.9\\
		GMFlow+~\cite{Xu2023UnifyingFlowStereo} & \underline{0.90} & 2.74 & 5.74 & 17.6\\
		MatchFlow~\cite{Dong2023RethinkingOpticalFlow} & 1.03 & 2.45 & 4.08 & 15.6\\
		CCMR+~\cite{Jahedi2024CCMRHighResolution} & 0.98 & 2.36 & - & 12.8\\
		RPKNet~\cite{Morimitsu2024RecurrentPartialKernel} & 1.12 & 2.45 & 3.78 & 13.0\\
            SEA-RAFT (L)~\cite{Wang2024SEARAFTSimple} & 1.19 & 4.11 & \underline{3.62} & 12.9\\
		FlowDiffuser~\cite{Luo2024FlowDiffuserAdvancingOptical} & \textbf{0.86} & \textbf{2.19} & - & \underline{11.8}\\
		DPFlow & 1.02 & \underline{2.26} & \textbf{3.37} & \textbf{11.1}\\
		\bottomrule
	\end{tabular}
	\label{tab:sintel_kitti_train}
\end{table}

We conduct ablation studies to evaluate the design choices of our dual-pyramid encoder.
We analyze the generalization results after stage 2 training with 100k iterations.

\subsubsection{Encoder Structure}

First, we assess the impact of our dual-pyramid structure and bidirectional recurrent unit.
\Cref{tab:ablation_structure} shows that both choices improve the results on both benchmarks.
The use of bidirectional GRU shows more significant improvements, highlighting the importance of bringing information from deeper features to shallower ones.

\subsubsection{Encoder Block}

Next, we check how the CGU compares to other commonly adopted network blocks.
\Cref{tab:ablation_block} shows that CGU achieves the best generalization results on almost all benchmarks.
In particular, we outperform the Swin Transformer~\cite{Liu2021SwinTransformerHierarchical} equipped with a cross-attention mechanism~\cite{Xu2022GMFlowLearningOptical}.
CGU also compares favorably against convolution-based approaches, such as ConvNeXt~\cite{Liu2022ConvNet2020s} and ResNet~\cite{He2016DeepResidualLearning}.

\subsubsection{Pyramid Depth}

DPFlow can adjust the pyramid depth at inference time to adapt to different inputs.
Here, we study how the depth change affects the results.
\Cref{tab:pyramid} shows the optimal number of levels is highly correlated to the input resolution, and the best results are usually obtained by adding another pyramid level when the input resolution doubles.
\section{Conclusions}
\label{sec:conclusions}

This paper addressed the current limitation of high-resolution optical flow estimation.
Our proposed DPFlow adopted a dual-pyramid framework with a recurrent network to produce high-quality optical flow estimation at multiple input resolutions in a single model.
Our experiments show that DPFlow achieves state-of-the-art results on multiple benchmarks, especially with high-resolution inputs.

\begin{table}[t]
    \caption{Ablation on the encoder structure.}
    \centering
    \begin{tabular}{@{}cccccc@{}}
        \toprule
      \multirow{2}{*}{Dual Pyr.} & \multirow{2}{*}{Bidir.} & \multicolumn{2}{c}{MPI-Sintel} & \multicolumn{2}{c}{KITTI 2015}\\
        \cmidrule(lr){3-4} \cmidrule(lr){5-6}
         & & clean & final & EPE & Fl-All\\
        \midrule
        $-$ & $-$ & 1.38 & 2.58 & 4.50 & 13.9\\
        $\checkmark$ & $-$ & 1.38 & 2.51 & 4.70 & 13.7\\
        $\checkmark$ & $\checkmark$ & \textbf{1.35} & \textbf{2.41} & \textbf{4.13} & \textbf{12.1}\\
        \bottomrule
    \end{tabular}
    \label{tab:ablation_structure}
\end{table}

\begin{table}[t]
    \caption{Ablation on the encoder block. FLOPs are calculated with a 1K-resolution input.}
    \centering
    \begin{tabular}{@{}lccccc@{}}
        \toprule
      \multirow{2}{*}{Encoder} & \multicolumn{2}{c}{MPI-Sintel} & \multicolumn{2}{c}{KITTI 2015} & FLOP\\
        \cmidrule(lr){2-3} \cmidrule(lr){4-5}
         & clean & final & EPE & Fl-All & (G)\\
        \midrule
        ResNet~\cite{He2016DeepResidualLearning} & 1.44 & 2.54 & 4.93 & 13.9 & 922\\
        ConvNeXt~\cite{Liu2022ConvNet2020s} & \textbf{1.33} & 2.50 & 4.34 & 13.3 & \textbf{674}\\
        Cross-Swin~\cite{Xu2022GMFlowLearningOptical} & 1.37 & 2.63 & 4.85 & 14.1 & 1187\\
        CGU & 1.35 & \textbf{2.41} & \textbf{4.13} & \textbf{12.1} & 778\\
        \bottomrule
    \end{tabular}
    \label{tab:ablation_block}
\end{table}

\begin{table}[t]
    \caption{Average results with different pyramid levels.}
    \centering
    \begin{tabular}{@{}llcccc@{}}
        \toprule
        \multirow{2}{*}{Dataset} & \multirow{2}{*}{Metric} & \multicolumn{4}{c}{Pyr. levels}\\
        \cmidrule(lr){3-6}
         & & 3 & 4 & 5 & 6\\
        \midrule
        KITTI 2015 (1K) & Fl-All $\downarrow$ & \textbf{11.1} & 11.7 & 11.7 & 12.7\\
        MPI-Sintel (1K) & EPE $\downarrow$ & \textbf{1.65} & 1.71 & 1.86 & 1.88\\
        Spring (2K) & 1px $\downarrow$ & 3.56 & \textbf{3.53} & 3.60 & 3.71\\
        VIPER (2K) & WAUC $\uparrow$ & \textbf{72.7} & 69.5 & 61.1 & 70.1\\
        Mbury-ST (3K) & EPE $\downarrow$ & 16.8 & 7.47 & \textbf{5.09} & 5.27\\
        \midrule
        Kubric-1K & \multirow{4}{*}{EPE $\downarrow$} & \textbf{0.50} & 0.53 & 0.59 & 0.61\\
        Kubric-2K &  & 0.92 & \textbf{0.85} & 0.86 & 1.04\\
        Kubric-4K &  & 2.57 & 1.80 & \textbf{1.77} & 1.81\\
        Kubric-8K &  & 14.7 & 5.97 & 4.37 & \textbf{4.11}\\
        \bottomrule
    \end{tabular}
    \label{tab:pyramid}
\end{table}

We also created the Kubric-NK dataset containing annotated optical flow samples for resolutions ranging from 1K to 8K.
This dataset provides a quantitative evaluation to compare the generalization capabilities to changes in input size.
Our high-resolution study provides new insights into designing more robust optical flow methods.
In particular, adaptive pyramids showed a significant advantage in counteracting the input size changes and providing stable predictions at all resolutions.
At high resolutions, DPFlow outperformed all previous methods by large margins.

\textbf{Limitations.}
DPFlow achieves comparable or superior results in almost all evaluations except the MPI-Sintel clean pass subset.
This subset contains synthetic images rendered without adverse conditions, such as motion blur and fog.
Therefore, other methods can still provide better accuracy when handling clean images at standard resolutions.
The proposed Kubric-NK dataset does not evaluate the effects of other conditions, such as lighting, fog, and fine details.
Combining adversary conditions with high-resolution annotated inputs remains an open topic.

\section{Acknowledgments}

This research is supported by 
the National Science Fund for Distinguished Young Scholars (62125601),
the Beijing Natural Science Foundation (IS23060),
the National Natural Science Foundation of China (62172035),
the Youth Teacher International Exchange \& Growth Program (No. QNXM20250001),
and the Fundamental Research Funds for the Central Universities (FRF-TP-2021-01C2, FRF-TP-22-048A1).
RMCJ is grateful to FAPESP (grants \#2022/15304-4), USP - UGPN, CNPq, CAPES, FINEP, and MCTI PPI-SOFTEX (TIC 13 DOU 01245.010222/2022-44).
{
    \small
    \bibliographystyle{ieeenat_fullname}
    \bibliography{aabib}
}

\clearpage
\maketitlesupplementary

\section{Training and evaluation details}

\Cref{tab:sup_hyperparams} shows the hyperparameters we used during each training stage.

\begin{table*}[t]
	\caption{Hyperparameters for each training stage. Batch sizes $\times 4$ indicate that gradients were accumulated during 4 iterations before backpropagating. Dataset abbreviations denote C: FlyingChairs~\cite{Dosovitskiy2015FlowNetlearningoptical}, T: FlyingThings3D~\cite{Mayer2016LargeDatasetTrain}, H: HD1K~\cite{Kondermann2016HCIBenchmarkSuite}, S: Sintel~\cite{Butler2012naturalisticopensource}, K: KITTI 2015~\cite{Menze2015Objectsceneflow}, Sp: Spring~\cite{Mehl2023SpringHighResolution}.}
	\centering
	\begin{tabular}{@{}ccccccc@{}}
		\toprule
		Stage & Datasets & Batch & Iters (Epochs) & LR & WD & GPU hrs.\\
		\midrule
		1 & C & 10 & 100k (45) & $2.5 \times 10^{-4}$ & $10^{-4}$ & 70\\
		2 & T & 6 & 1M (80) & $1.25 \times 10^{-4}$ & $10^{-4}$ & 700\\
		3 & T+H+S+K & $6 \times 4$ & 120k (25) & $1.25 \times 10^{-4}$ & $10^{-5}$ & 300\\
		4a & K & 6 & 2k (250) & $1 \times 10^{-4}$ & $10^{-5}$ & 6\\
		4b & Sp & $4 \times 4$ & 120k (100) & $1 \times 10^{-4}$ & $10^{-5}$ & 360\\
		\bottomrule
	\end{tabular}
	\label{tab:sup_hyperparams}
\end{table*}

\section{Checkpoint results}

We report the results of our model after each of the four training stages in~\cref{tab:sup_stage_results}.
These results can be used as a sanity check to reproduce the training of our model.

\begin{table}[t]
	\caption{Results after each training stage. Values in parentheses indicate that the same dataset was used during training.}
	\centering
	\begin{tabular}{@{}lcccc@{}}
		\toprule
		\multirow{2}{*}{Stage} & S. Clean & S. Final & \multicolumn{2}{c}{KITTI 2015}\\
            \cmidrule(lr){4-5}
		  & EPE & EPE & EPE & Fl-All\\
		\midrule
		1 & 2.74 & 3.91 & 11.1 & 32.1\\
		2 & 1.02 & 2.26 & 3.37 & 11.1\\
		3 & (0.55) & (0.86) & (1.04) & (3.18)\\
		4a & 0.80 & 1.24 & (0.83) & (2.24)\\
		4b & 1.37 & 2.45 & 6.36 & 15.7\\
		\bottomrule
	\end{tabular}
	\label{tab:sup_stage_results}
\end{table}

\section{Model details}

\Cref{tab:sup_model_details} details different models' size, time, and memory costs with input resolutions ranging from 1K to 8K.
Due to deep feature pyramids, DPFlow has low computational costs and offers relatively low latency, being faster than most top-performing methods.
FlowFormer++ and MatchFlow can reduce memory consumption using input tiling, but this strategy increases inference time considerably.
RPKNet can achieve lower computational costs, but DPFlow offers a $23\%$ improvement on the KITTI 2015 benchmark.

\begin{table*}[t]
    \caption{Model details. $^{\Box}$ denotes methods that use input tiling~\cite{Jaegle2022PerceiverIOgeneral,Huang2022FlowFormerTransformerArchitecture}. $^*$ indicates that the model is running in local correlation mode, which decreases memory consumption but increases inference time. $\dag$ results are not available due to running out of memory. Times are calculated using an NVIDIA RTX3090 GPU.}
    \centering
    \begin{tabular}{@{}lcccccccccc@{}}
        \toprule
        \multirow{2}{*}{Model} & KITTI 2015 & Params & \multicolumn{4}{c}{Time (s)} & \multicolumn{4}{c}{Memory (GB)}\\
        \cmidrule(lr){4-7} \cmidrule(lr){8-11}
         & (Fl-All) & (M) & 1K & 2K & 4K & 8K & 1K & 2K & 4K & 8K\\
        \midrule
        RAFT~\cite{Teed2020RAFTrecurrentall}$^*$ & 5.10 & \underline{5.25} & 0.51 & 1.36 & 4.76 & $\dag$ & \underline{0.65} & 1.71 & 6.40 & $\dag$\\
        GMA~\cite{Jiang2021LearningEstimateHidden} & 5.25 & 5.88 & 0.18 & 1.17 & $\dag$ & $\dag$ & 1.41 & 17.3 & $\dag$ & $\dag$\\ 
        DIP~\cite{Zheng2022DIPDeepInverse} & 4.21 & 5.37 & 0.51 & 2.11 & 8.60 & $\dag$ & 1.59 & 5.46 & 20.9 & $\dag$\\
        SKFlow~\cite{Sun2022SKFlowLearningOptical} & 4.84 & 6.27 & 0.31 & 1.70 & $\dag$ & $\dag$ & 1.17 & 17.4 & $\dag$ & $\dag$\\
        FlowFormer++~\cite{Shi2023FlowFormerMaskedCost}$^{\Box}$ & 4.52 & 16.1 & 1.17 & 2.68 & 9.02 & 29.9 & 4.24 & 4.37 & 6.10 & \underline{18.7}\\
        MatchFlow~\cite{Dong2023RethinkingOpticalFlow}$^{\Box}$ & 4.63 & 15.4 & 0.63 & 1.44 & 5.82 & 22.8 & 1.31 & \underline{1.61} & \textbf{3.04} & 20.7\\
        MS-RAFT+~\cite{Jahedi2023MSRAFTHigh}$^*$ & 4.19 & 16.1 & 0.74 & 2.82 & $\dag$ & $\dag$ & 2.61 & 9.34 & $\dag$ & $\dag$\\
        GMFlow+~\cite{Xu2023UnifyingFlowStereo} & 4.49 & 7.36 & 0.23 & 1.69 & $\dag$ & $\dag$ & 2.99 & 20.6 & $\dag$ & $\dag$\\
        RPKNet~\cite{Morimitsu2024RecurrentPartialKernel}$^*$ & 4.64 & \textbf{2.84} & \textbf{0.08} & \textbf{0.28} & \textbf{1.18} & \textbf{5.34} & \textbf{0.56} & \textbf{1.34} & \underline{4.42} & \textbf{16.8}\\
        SEA-RAFT (L)~\cite{Wang2024SEARAFTSimple}$^*$ & 4.30 & 19.6 & 0.24 & 0.70 & 2.60 & $\dag$ & 0.79 & 2.02 & 7.21 & $\dag$\\
        CCMR+~\cite{Jahedi2024CCMRHighResolution}$^*$ & 3.86 & 11.5 & 1.34 & 5.36 & $\dag$ & $\dag$ & 3.19 & 12.1 & $\dag$ & $\dag$\\
        DPFlow$^*$ & \textbf{3.56} & 10.0 & \underline{0.16} & \underline{0.53} & \underline{1.98} & \underline{8.40} & 0.76 & 1.90 & 6.70 & 20.7\\
        \bottomrule
    \end{tabular}
    \label{tab:sup_model_details}
\end{table*}

\section{Evaluation details}

\subsection{Metrics}

This section explains how to calculate the metrics we use for the evaluations in each benchmark.
Let $F \in \mathbb{R}^{N \times 2}$ be the predicted flow and $G \in \mathbb{R}^{N \times 2}$ the groundtruth, where $N = HW$ for height $H$ and width $W$.
Then, the metrics are calculated as follows.

\textbf{End-Point-Error (EPE):} consists of the average Euclidean distance between the predicted flow vectors and the groundtruth ones:
\begin{equation}
    \text{EPE}(F, G) = \frac{1}{N} \sum_{i=1}^{N}{\|F_i - G_i\|_2}.
\end{equation}

\textbf{Outlier ratio:} represents the percentage of predictions whose Euclidean distances are beyond a given threshold $\tau$:
\begin{equation}
    \text{Outlier}(F, G) = \frac{1}{N} \sum_{i=1}^{N}{\left[
    \|F_i - G_i\|_2 > \tau \right]},
\end{equation}
where $[\cdot]$ is the Iverson bracket.
Each benchmark selects a specific threshold value and represents its metric by a different name.
For example, the KITTI benchmark~\cite{Menze2015Objectsceneflow} adopts $\tau = \max{(3, 0.05\|G_i\|_2)}$ and calls their metric \textbf{Fl-All}.
On the other hand, the Spring benchmark~\cite{Mehl2023SpringHighResolution} uses $\tau = 1$ and names their metric as \textbf{1px}.

\textbf{Weighted Area Under the Curve (WAUC):} used by the VIPER dataset~\cite{Richter2017PlayingBenchmarks} to integrate over multiple thresholds and give more importance to more accurate results.
\begin{equation}
    \begin{aligned}
        \text{WAUC}&(F, G) =\\
         &\frac{100}{N \sum_{i=1}^{100} w_i} \sum_{i=1}^{100} w_i \sum_{j=1}^{N} [\|F_j - G_j\|_2 \leq \delta_i],
    \end{aligned}
\end{equation}
where $[\cdot]$ is the Iverson bracket, $\delta_i = \frac{i}{20}$ and $w_i = 1 - \frac{i - 1}{100}$.

\subsection{Spring dataset metrics}

To improve the evaluation of predictions for very thin structures, the Spring benchmark~\cite{Mehl2023SpringHighResolution} proposed to generate the groundtruth at doubled resolution (image is 2K and groundtruth is 4K).
Therefore, each prediction (at 2K) is compared with its four corresponding groundtruth values at 4K, and the lowest error is chosen as the metric.
We follow the same procedure when evaluating the train results at 2K in Tab. 3 in the main paper.
Unlike the study by SEA-RAFT~\cite{Wang2024SEARAFTSimple}, we also use the images at full resolution instead of downsampling them to 1K, which explains the difference in our results.

For the Spring train 4K evaluation in Tab. 3 in the main paper, we use bicubic interpolation to upsample the images from 2K to 4K.
We also double the flow groundtruth values since they are originally encoded for a 2K resolution evaluation.
Since the upsampling may create artifacts at the motion boundaries, we create a mask to ignore pixels near the borders.
We create this mask using a strategy similar to the one the authors proposed for handling thin structures.
We first reduce the 4K groundtruth to 2K resolution by stacking $2 \times 2$ pixel blocks.
Then, we calculate the EPE metric between all combinations of the four pixels inside each $2 \times 2$ block.
If the maximum EPE within a block is greater than one, we consider this block to lie on a motion boundary and ignore all four pixels.
The Python code describing this procedure is shown in~\cref{lst:sup_spring_4k_eval}.

\begin{lstlisting}[float=t, basicstyle=\footnotesize, language=Python, showstringspaces=false, label=lst:sup_spring_4k_eval, caption={Python code for masking out pixels on the motion boundaries to calculate the metric for the Spring train 4K evaluation in Tab. 3 in the main paper.}]
import torch
import torch.nn.functional as F
from einops import rearrange

def compute_valid_mask(flow: torch.Tensor):
    # flow shape is [B, 2, H, W]
    # downsample by stacking 2x2 blocks
    flow_stack = rearrange(
        flow,
        "b c (h nh) (w nw) -> b (nh nw) c h w",
        nh=2,
        nw=2
    )
    # calculate EPE between all pairs
    # in each 2x2 block
    flow_stack4 = flow_stack.repeat(
        1, 4, 1, 1, 1)
    flow_stack4 = rearrange(
        flow_stack4,
        "b (m n) c h w -> b m n c h w",
        m=4)
    diff = flow_stack[:, :, None] - flow_stack4
    diff = rearrange(
        diff,
        "b m n c h w -> b (m n) c h w"
    )
    diff = torch.sqrt(torch.pow(diff, 2).sum(2))
    # mark pixels as valid only if the maximum
    # EPE of their block is below one
    max_diff, _ = diff.max(1)
    max_diff = F.interpolate(
        max_diff[:, None],
        scale_factor=2,
        mode="nearest"
    )
    valid_mask = max_diff < 1.0
    return valid_mask
\end{lstlisting}

\section{Choosing the evaluation checkpoint}

Here, we include the individual results for each method we used to calculate the averages presented in Tab. 2 in the main paper.

\subsection{Spring dataset}

Due to the large size of the Spring dataset~\cite{Mehl2023SpringHighResolution}, we only use the first $10\%$ samples from each sequence for this experiment (480 samples).
The results are presented in~\cref{tab:sup_ckpt_comparison_spring}.

\begin{table}[t]
    \caption{Training stage experiments on the Spring dataset.}
    \centering
    \begin{tabular}{@{}lcc@{}}
        \toprule
        \multirow{2}{*}{Method} & \multicolumn{2}{c}{1px $\downarrow$}\\
        \cmidrule(lr){2-3}
         & Stage 2 & Stage 3\\
        \midrule
        PWC-Net & 7.18 & 5.15\\
        IRR & 5.46 & 4.52\\
        VCN & 7.53 & 8.08\\
        RAFT & 4.45 & 3.85\\
        GMA & 4.33 & 3.67\\
        DIP & 4.36 & 4.21\\
        GMFlow & 5.92 & 6.83\\
        FlowFormer & 4.16 & 3.93\\
        SKFlow & 4.61 & 3.93\\
        FlowFormer++ & 4.1 & 4.21\\
        MatchFlow & 4.33 & 4.15\\
        GMFlow+ & 5.04 & 5.07\\
        RPKNet & 3.95 & 3.86\\
        SEA-RAFT (M) & 4.37 & 3.63\\
        RAPIDFlow & 4.32 & 3.94\\
        DPFlow & 4.76 & 3.53\\
        \midrule
        Average & 4.92 & \textbf{4.53}\\
        \bottomrule
    \end{tabular}
    \label{tab:sup_ckpt_comparison_spring}
\end{table}

\subsection{VIPER dataset}

Due to the large size of the VIPER dataset~\cite{Richter2017PlayingBenchmarks}, we only use the first $10\%$ samples from each sequence for this experiment (473 samples).
The results are presented in~\cref{tab:sup_ckpt_comparison_viper}.

\begin{table}[t]
    \caption{Training stage experiments on the VIPER dataset.}
    \centering
    \begin{tabular}{@{}lcc@{}}
        \toprule
        \multirow{2}{*}{Method} & \multicolumn{2}{c}{WAUC $\uparrow$}\\
        \cmidrule(lr){2-3}
         & Stage 2 & Stage 3\\
        \midrule
        PWC-Net & 51.2 & 61.3\\
        IRR & 58.1 & 61.4\\
        VCN & 55.6 & 58.4\\
        RAFT & 63.7 & 67.1\\
        GMA & 64 & 67.7\\
        DIP & 66.8 & 67.7\\
        GMFlow & 52.9 & 52.4\\
        FlowFormer & 66.1 & 68.8\\
        SKFlow & 64.4 & 67.8\\
        FlowFormer++ & 66.8 & 68.6\\
        MatchFlow & 66.3 & 69.0\\
        GMFlow+ & 64.8 & 65.8\\
        RPKNet & 66.9 & 68.7\\
        SEA-RAFT (M) & 68.9 & 71.8\\
        RAPIDFlow & 62.7 & 66.5\\
        DPFlow & 70.4 & 72.7\\
        \midrule
        Average & 63.1 & \textbf{65.9}\\
        \bottomrule
    \end{tabular}
    \label{tab:sup_ckpt_comparison_viper}
\end{table}

\subsection{Middlebury-ST dataset}

We use all 23 training samples of the Middlebury-ST dataset~\cite{Scharstein2014Highresolutionstereo} for this experiment.
The results are presented in~\cref{tab:sup_ckpt_comparison_mbury}.

\begin{table}[t]
    \caption{Training stage experiments on the Middlebury-ST dataset.}
    \centering
    \begin{tabular}{@{}lcc@{}}
        \toprule
        \multirow{2}{*}{Method} & \multicolumn{2}{c}{EPE $\downarrow$}\\
        \cmidrule(lr){2-3}
         & Stage 2 & Stage 3\\
        \midrule
        PWC-Net & 35.9 & 74.3\\
        IRR & 15.2 & 61.4\\
        VCN & 44.2 & 51\\
        RAFT & 35.2 & 51\\
        DIP & 16.7 & 21.2\\
        FlowFormer & 19.0 & 13.8\\
        FlowFormer++ & 17.5 & 15.2\\
        MatchFlow & 16.7 & 19.4\\
        RPKNet & 15.0 & 16.4\\
        SEA-RAFT (M) & 65.1 & 59.2\\
        RAPIDFlow & 6.41 & 7.33\\
        DPFlow & 4.82 & 5.25\\
        \midrule
        Average & \textbf{24.3} & 32.9\\
        \bottomrule
    \end{tabular}
    \label{tab:sup_ckpt_comparison_mbury}
\end{table}

\subsection{Kubric-1K dataset}

We use all 600 samples at 1K resolution from our proposed Kubric-NK dataset for this experiment.
The results are presented in~\cref{tab:sup_ckpt_comparison_kubric}.

\begin{table}[t]
    \caption{Training stage experiments on the Kubric-1K dataset.}
    \centering
    \begin{tabular}{@{}lcc@{}}
        \toprule
        \multirow{2}{*}{Method} & \multicolumn{2}{c}{EPE $\downarrow$}\\
        \cmidrule(lr){2-3}
         & Stage 2 & Stage 3\\
        \midrule
        PWC & 1.4 & 1.28\\
        IRR & 0.95 & 1.1\\
        VCN & 1.23 & 1.09\\
        RAFT & 0.73 & 0.69\\
        GMA & 0.7 & 0.67\\
        DIP & 0.71 & 0.68\\
        GMFlow & 0.68 & 0.88\\
        FlowFormer & 0.5 & 0.52\\
        SKFlow & 0.69 & 0.62\\
        FlowFormer++ & 0.48 & 0.46\\
        MatchFlow & 0.54 & 0.54\\
        GMFlow+ & 0.48 & 0.45\\
        MemFlow & 0.57 & 0.55\\
        SEA-RAFT(M) & 0.56 & 0.53\\
        RPKNet & 0.61 & 0.6\\
        RAPIDFlow & 0.85 & 0.81\\
        DPFlow & 0.57 & 0.5\\
        \midrule
        Average & 0.72 & \textbf{0.70}\\
        \bottomrule
    \end{tabular}
    \label{tab:sup_ckpt_comparison_kubric}
\end{table}

\section{Feature visualization}

\begin{figure*}
    \centering
    \includegraphics[width=\textwidth]{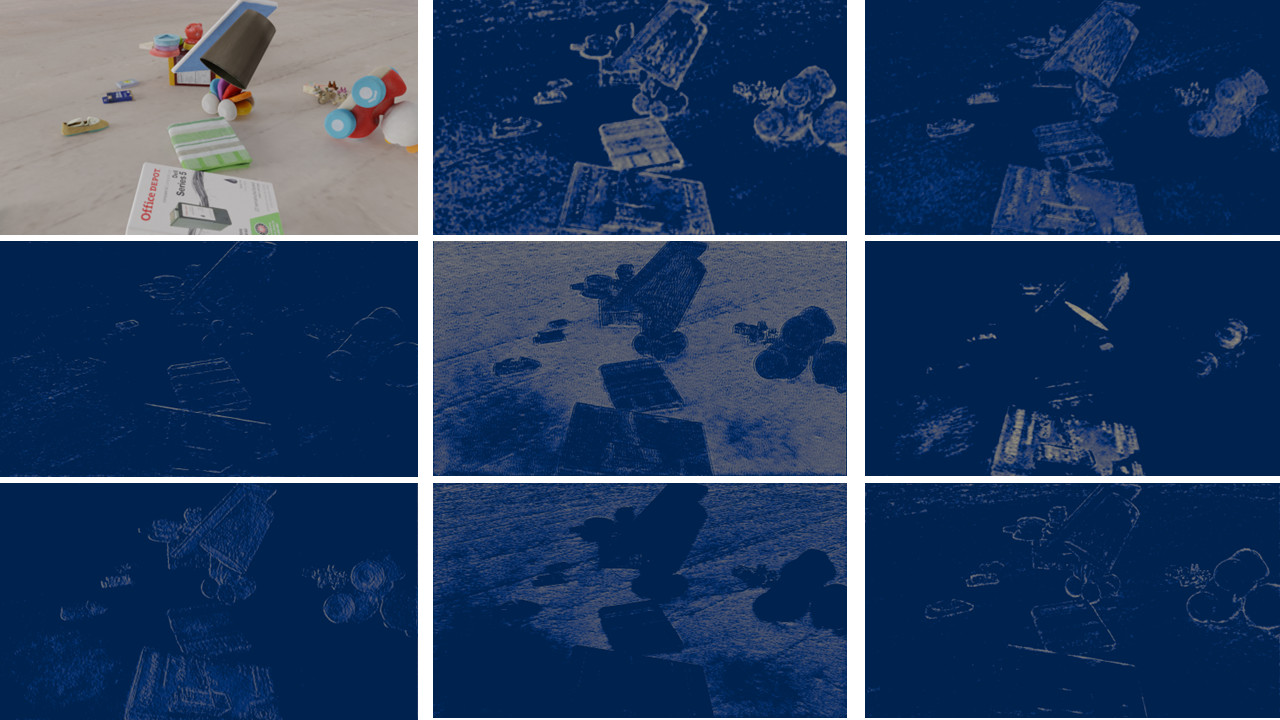}
    \caption{Visualization of cross-gate activations on an 8K sample from the Kubric-NK dataset.}
    \label{fig:sup_feature_viz_kubric}
\end{figure*}

\begin{figure*}
    \centering
    \includegraphics[width=\textwidth]{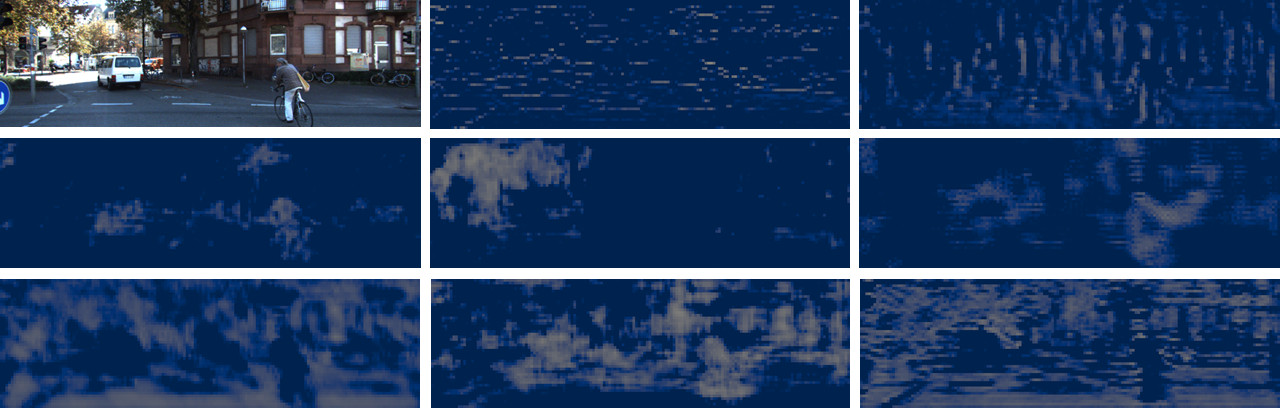}
    \caption{Visualization of cross-gate activations on a 1K sample from the KITTI 2015 dataset.}
    \label{fig:sup_feature_viz_kitti}    
\end{figure*}

We study the activation patterns of the CGU cross-gate to see what type of pattern they focus on.
Here, we select the output feature of the encoder network and save the gate activations using heatmaps.
Figures~\ref{fig:sup_feature_viz_kubric} and~\ref{fig:sup_feature_viz_kitti} show the gate activations on samples from the Kubric-NK (8K resolution) and KITTI 2015 (1K), respectively.
We observe that some gates can capture some structural information about the scene.
For example, some channels focus on horizontal and vertical lines, while others separate background from foreground and encode motion boundaries.

\section{Input downsampling}

One popular strategy to avoid handling high-resolution inputs consists of first downsampling the inputs and then upsampling the predictions back to the original resolution.
Here, we conduct a study to evaluate how this input downsampling affects the results.
We use samples from three high-resolution datasets: Spring~\cite{Mehl2023SpringHighResolution} (2K), Middlebury-ST~\cite{Scharstein2014Highresolutionstereo} (3K), and Kubric-NK (8K).
For this study, we use bilinear interpolation to first downsample the inputs to 1K resolution, and then upsample the predictions back to the original sizes before computing the accuracy metrics.
The results in~\cref{tab:sup_input_downsampling} show some interesting observations.
Downsampling from 2K to 1K in the Spring benchmark negatively impacts all methods except GMFlow+.
This result indicates that most approaches are robust to this input size, but the global matching strategy from GMFlow+ may have more bias towards the training sizes.
On the Middlebury-ST 3K samples, we observe significant improvements in methods with fixed structures.
On the other hand, adaptable methods like DPFlow and RAPIDFlow still benefit from processing the images at the original resolution.
Using an eight-times downsampling on Kubric-NK benefits almost all methods.
However, these results may be caused by the relatively simple shapes and motions generated by Kubric~\cite{Greff2022Kubricscalabledataset}.
Since the objects do not present fine-grained details, downsampling them does not affect the input contents significantly.
A more thorough study of the effects of downsampling would require a high-resolution dataset with finer and more complex motion patterns.

\begin{table*}[t]
	\caption{Results on high-resolution benchmarks with and without downsampling the inputs. The @1K columns indicate that the inputs are downsampled to 1K resolution and the predictions are upsampled back to the original sizes. $\dag$ results are not available due to running out of memory.}
	\centering
	\begin{tabular}{@{}lcccccc@{}}
		\toprule
		  \multirow{2}{*}{Method} & \multicolumn{2}{c}{Spring (1px $\downarrow$)} & \multicolumn{2}{c}{Mbury-ST (EPE $\downarrow$)} & \multicolumn{2}{c}{Kubric-NK (EPE $\downarrow$)}\\
		\cmidrule(lr){2-3} \cmidrule(lr){4-5} \cmidrule(lr){6-7}
		  & 2K & @1K & 3K & @1K & 8K & @1K\\
		\midrule
            RAFT~\cite{Teed2020RAFTrecurrentall} & 3.85 & 4.52 & 35.2 & 8.01 & 82.7 & 5.09\\
            GMA~\cite{Jiang2021LearningEstimateHidden} & 3.75 & 4.53 & $\dag$ & 8.23 & $\dag$ & 4.91\\ 
            DIP~\cite{Zheng2022DIPDeepInverse} & 3.64 & 3.87 & 16.7 & 8.01 & $\dag$ & 5.42\\
            SKFlow~\cite{Sun2022SKFlowLearningOptical} & 3.79 & 4.20 & $\dag$ & 6.88 & $\dag$ & 4.72\\ 
            FlowFormer++~\cite{Shi2023FlowFormerMaskedCost} & 3.76 & 4.21 & 17.5 & 6.93 & 24.4 & \textbf{3.39}\\ 
            MatchFlow~\cite{Dong2023RethinkingOpticalFlow} & 3.97 & 4.21 & 16.7 & 7.18 & 29.0 & 3.89\\ 
            GMFlow+~\cite{Xu2023UnifyingFlowStereo} & 5.72 & 4.00 & $\dag$ & \underline{5.67} & $\dag$ & 4.43\\ 
            RPKNet~\cite{Morimitsu2024RecurrentPartialKernel} & \underline{3.28} & 4.12 & 15.0 & 6.74 & 18.2 & 4.59\\ 
            SEA-RAFT (M)~\cite{Wang2024SEARAFTSimple} & 3.47 & \textbf{3.53} & 65.1 & 6.71 & 51.8 & 4.07\\
            RAPIDFlow~\cite{Morimitsu2024RAPIDFlowRecurrentAdaptable} & 3.73 & 5.06 & \underline{6.41} & 8.99 & \underline{5.96} & 5.99\\
            DPFlow & \textbf{3.06} & \underline{3.57} & \textbf{5.09} & \textbf{5.63} & \textbf{4.13} & \underline{3.88}\\
		\bottomrule
        \end{tabular}
	\label{tab:sup_input_downsampling}
\end{table*}

\section{Kubric-NK}

This section details the Kubric-NK dataset and presents additional discussions about the effect of input resolution on the performance of optical flow methods.

\subsection{Dataset samples}

\begin{figure*}
    \centering
    \includegraphics[height=0.95\textheight]{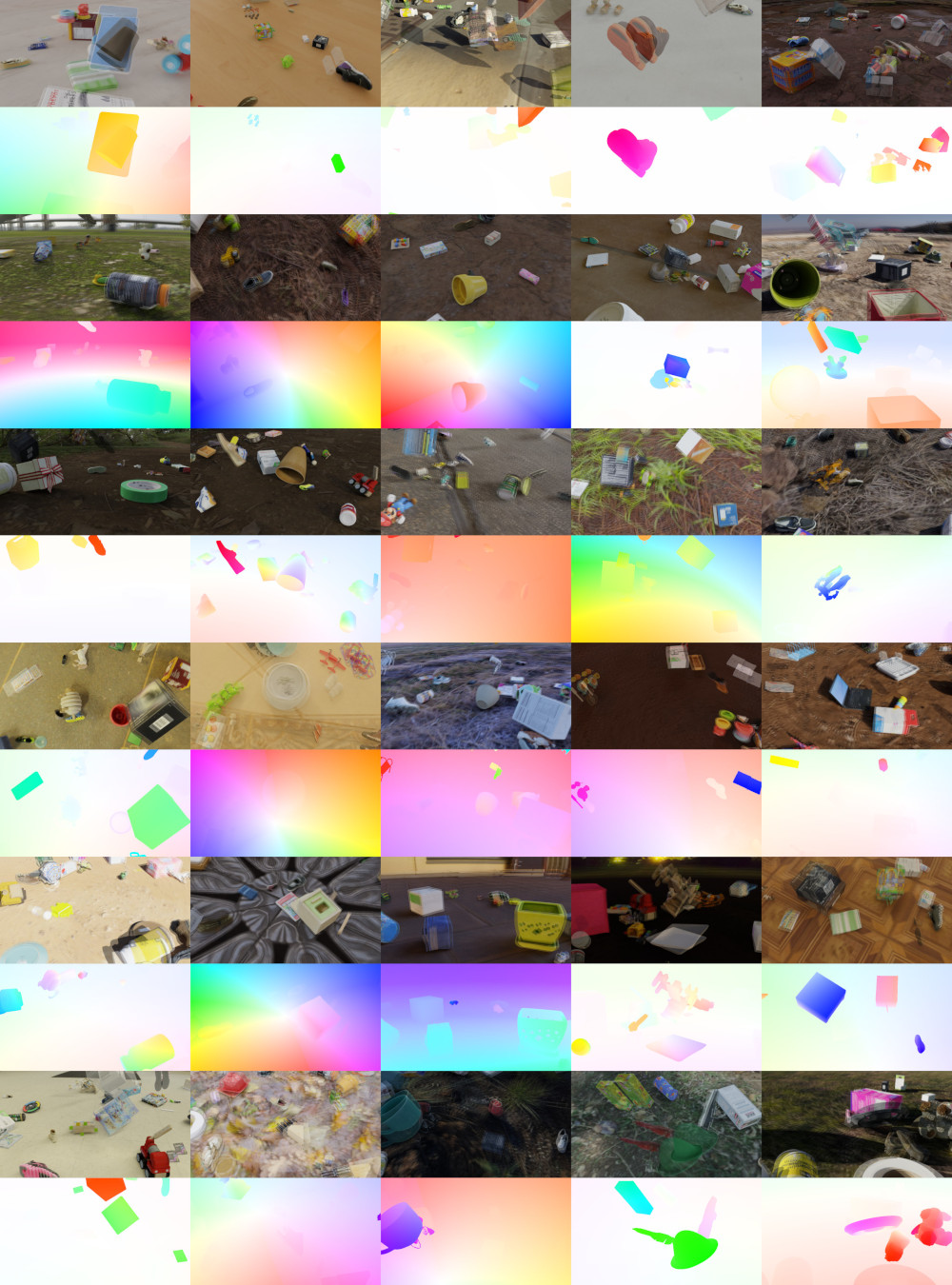}
    \caption{Samples from all 30 sequences from the Kubric-NK dataset. Odd rows show two consecutive images overlayed; even rows show their respective flow.}
    \label{fig:sup_kubric_nk_collage}
\end{figure*}

\Cref{fig:sup_kubric_nk_collage} shows an image pair and its respective optical flow annotation for each of the 30 sequences of the Kubric-NK dataset.

\subsection{Generalization to different resolutions}

\begin{figure}
    \centering
    \begin{subfigure}{\linewidth}
        \centering
        \includegraphics[width=\linewidth]{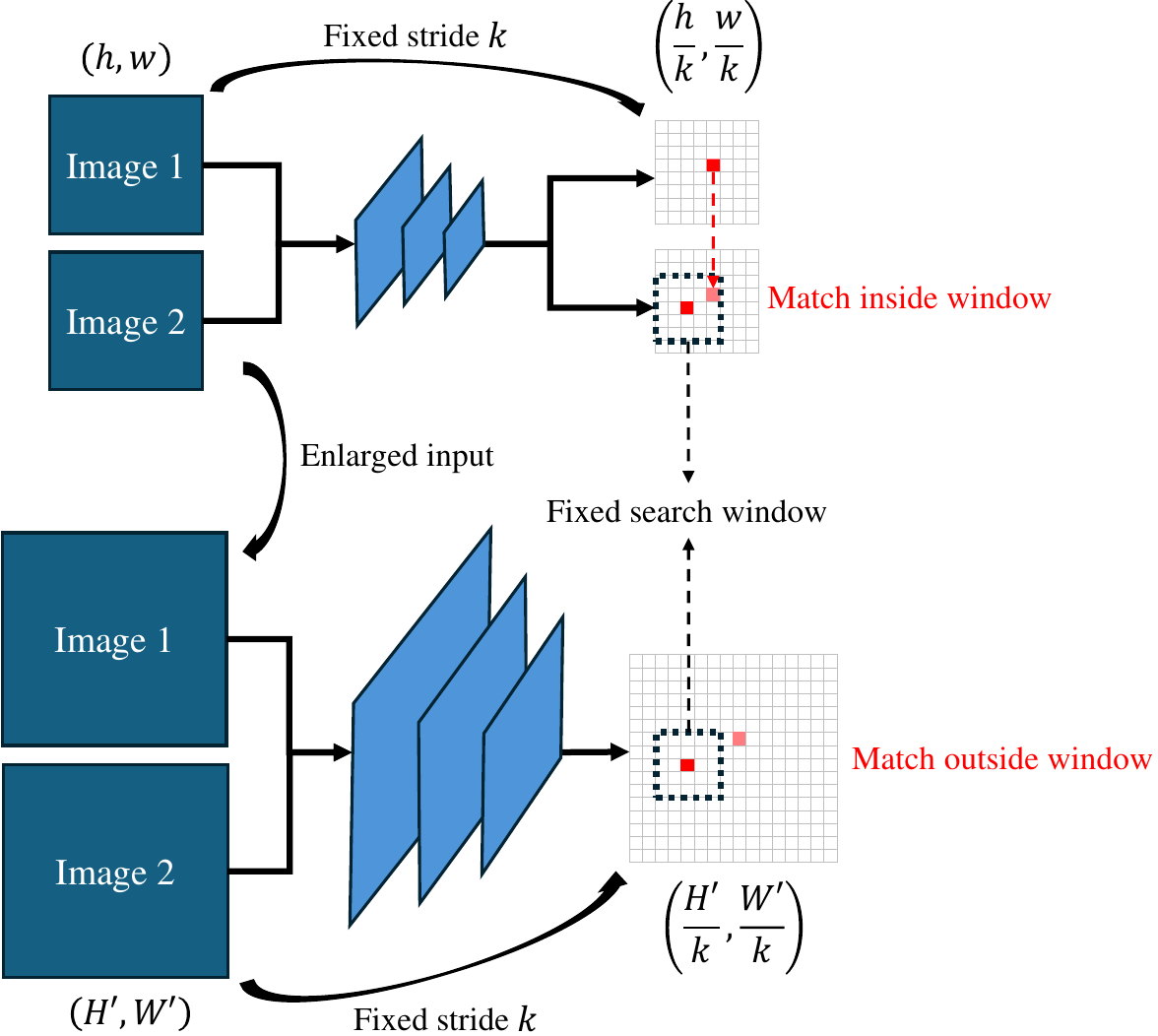}
        \caption{Local search with fixed encoder}
        \label{fig:sup_fixed_encoder}
    \end{subfigure}
    \\
    \begin{subfigure}{\linewidth}
        \centering
        \includegraphics[width=\linewidth]{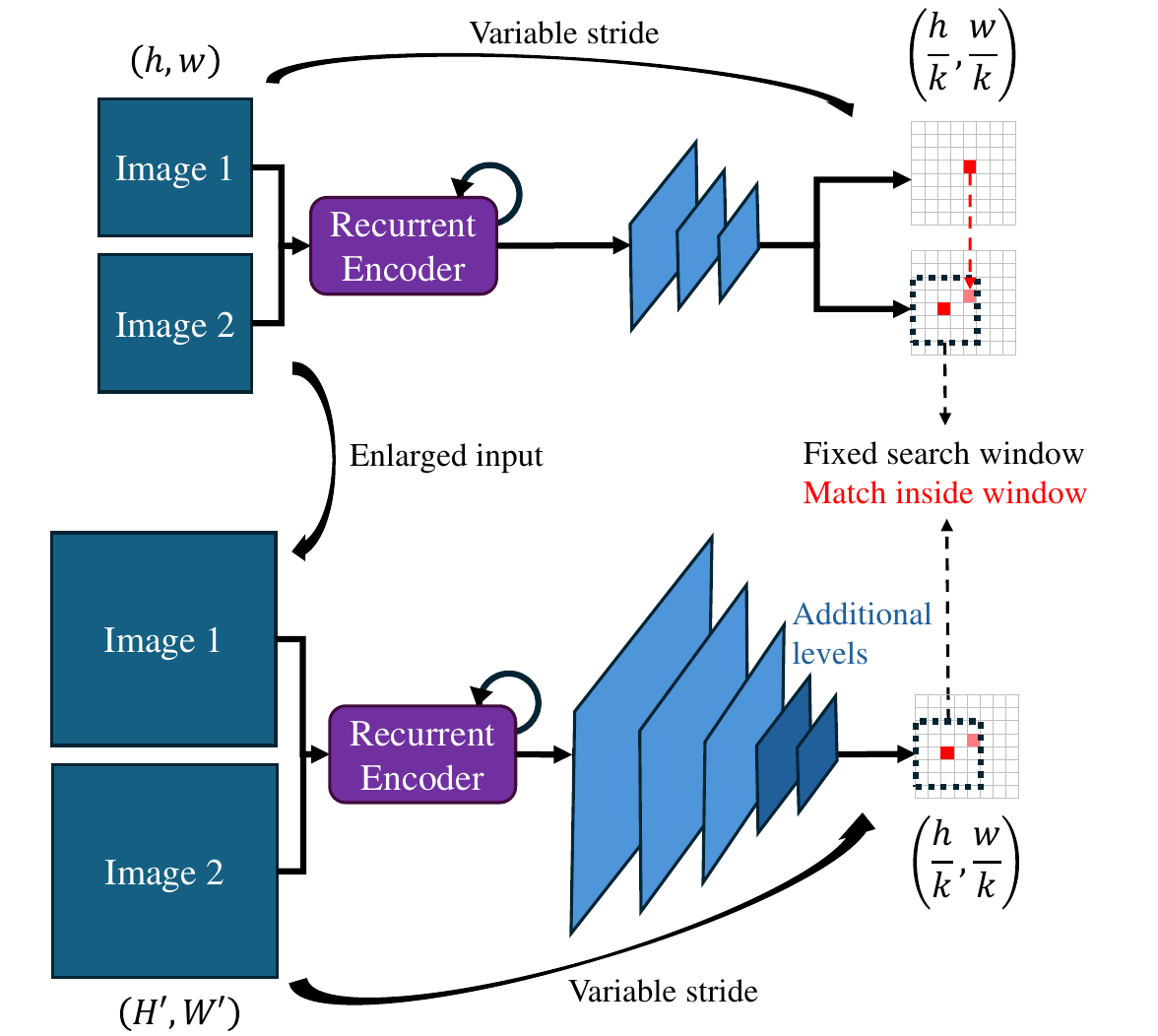}
        \caption{DPFlow's adaptable encoder}
        \label{fig:sup_recurrent_encoder}
    \end{subfigure}
    \caption{Simple example of the local matching procedure on typical optical flow models. (a) The spatial size of the feature map extracted by an encoder with a fixed structure is determined by the encoder's total stride $k$ (\eg, 8 times smaller than the input). When the input resolution increases, the matched points are beyond the local window's range, causing a mismatch. (b) DPFlow adopts a recurrent encoder that can be dynamically unfolded to change the number of encoding levels and control the size of the initial feature map. This alleviates the window range problem.}
    \label{fig:sup_problem_large_resolution}
\end{figure}

Optical flow estimation is fundamentally a point-matching problem between a pair of images.
Therefore, the overall magnitude of the flow (correspondence) vectors tends to correlate with the size of the inputs, making it more difficult for higher-resolution inputs.
There are two main optical flow estimation approaches: global and local matching.
Global matching can theoretically handle changes in the input size and large motions.
However, the cost of global search increases quadratically according to the input size, which makes it unfeasible for larger inputs.
Local matching only searches for correspondences within a limited window (typically $9 \times 9$) around the origin point.
This approach, however, requires the corresponding points to be within the range of the search window.
This causes a problem for most local methods because they adopt an encoder with a fixed structure, which makes the size of the feature directly determined by the input resolution.
\Cref{fig:sup_problem_large_resolution} illustrates how the increase in input size affects the matching procedure and how DPFlow's recurrent encoder alleviates this problem.

\subsubsection{Training bias}

Besides the problems caused by the limited search range, the training distribution also heavily affects the performance of non-adaptive models to different inputs.
We demonstrate the effect of training bias with two studies based on the open-source models Flow1D~\cite{Xu2021HighResolutionOptical} and CroCo~\cite{Revaud2023CroCov2Improved}.

Flow1D provides an additional model variant trained using a dataset with high-resolution samples.
We compare the performance of Flow1D on Kubric-NK using the standard (trained for MPI-Sintel) and high-resolution training in~\cref{tab:sup_flow1d_results}.
The results demonstrate that high-resolution training is effective and necessary to produce stable predictions at 4K resolution.
However, we also observe that it negatively affects the results at the lower 1K resolution, thus clearly showing the effect of the training bias.

\begin{table}[t]
    \caption{Flow1D results using standard-resolution stage 3 training (for MPI-Sintel benchmark) and after high-resolution re-training.}
    \centering
    \begin{tabular}{@{}lccc@{}}
        \toprule
        \multirow{2}{*}{Train resolution} & \multicolumn{3}{c}{Kubric-NK (EPE $\downarrow$)}\\
        \cmidrule(lr){2-4}
         & 1K & 2K & 4K\\
        \midrule
        Standard (Sintel) & 0.85 & 1.80 & 32.7\\
        High-res. & 0.88 & 1.42 & 4.49\\
        \bottomrule
    \end{tabular}
    \label{tab:sup_flow1d_results}
\end{table}

Our second study evaluates the performance of the CroCo model on the training split of the Middlebury v3 stereo-matching benchmark~\cite{Scharstein2014Highresolutionstereo}.
This benchmark contains 15 image pairs ranging from 1K to 3K resolution inputs.
Since stereo-matching can be viewed as a subset of optical flow (matching only on the horizontal axis), optical flow models can be used for this task by zeroing the y-axis prediction.
We chose the CroCo model because it provides trained variants for stereo and optical flow tasks using a common backbone network.
The only differences between these variants are the number of output prediction heads (1 for stereo vs. 2 for optical flow) and the training samples.

\Cref{tab:sup_croco_results} shows how using these two variants affects the Middlebury v3 training benchmark results.
Although optical flow and stereo matching are both similar matching problems, directly transferring the optical flow-trained model for stereo estimation results in a noticeable degradation.
Since both variants share the same architecture, these results further illustrate the effect of the training bias.

\begin{table}[t]
    \caption{Results of evaluating CroCo~\cite{Revaud2023CroCov2Improved} variants trained on stereo and optical flow samples on the Middlebury v3 stereo matching benchmark.}
    \centering
    \begin{tabular}{@{}lccc@{}}
        \toprule
        \multirow{2}{*}{Variant} & \multicolumn{3}{c}{Middlebury v3 (training)}\\
        \cmidrule(lr){2-4}
         & EPE $\downarrow$ & 1px $\downarrow$ & 2px $\downarrow$\\
        \midrule
        CroCo-Stereo & 0.37 & 3.86 & 1.93\\
        CroCo-Flow & 7.17 & 29.7 & 20.1\\
        \bottomrule
    \end{tabular}
    \label{tab:sup_croco_results}
\end{table}

These studies demonstrated that the problems of handling different inputs can be mitigated with specialized training.
In particular, training different variants for particular input resolutions can result in significant improvements.
Nonetheless, this strategy is only palliative since it can only shift the problem to another resolution range.
Moreover, training multiple variants of each model is expensive, especially with high-resolution inputs.
One advantage of the proposed DPFlow is that it significantly reduces the input resolution gap by changing its pyramids on the fly without requiring additional training.

\subsection{Additional qualitative results}

\begin{figure*}
    \centering
    \includegraphics[width=\linewidth]{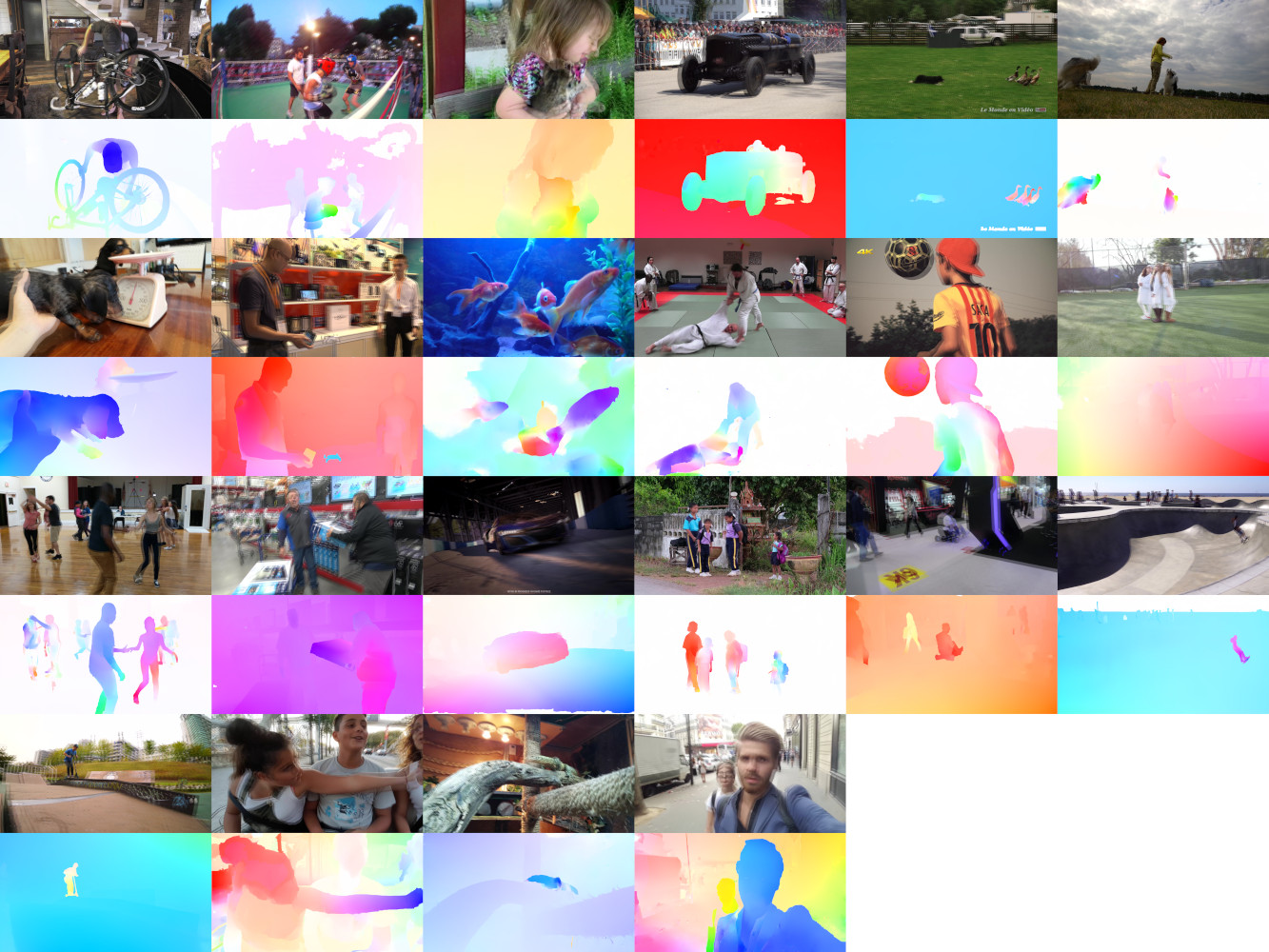}
    \caption{Qualitative results on 4K samples from the DAVIS dataset~\cite{PontTuset20172017DAVISChallenge}.}
    \label{fig:sup_davis_qualitative}
\end{figure*}

\begin{figure*}
    \centering
    \includegraphics[width=\linewidth]{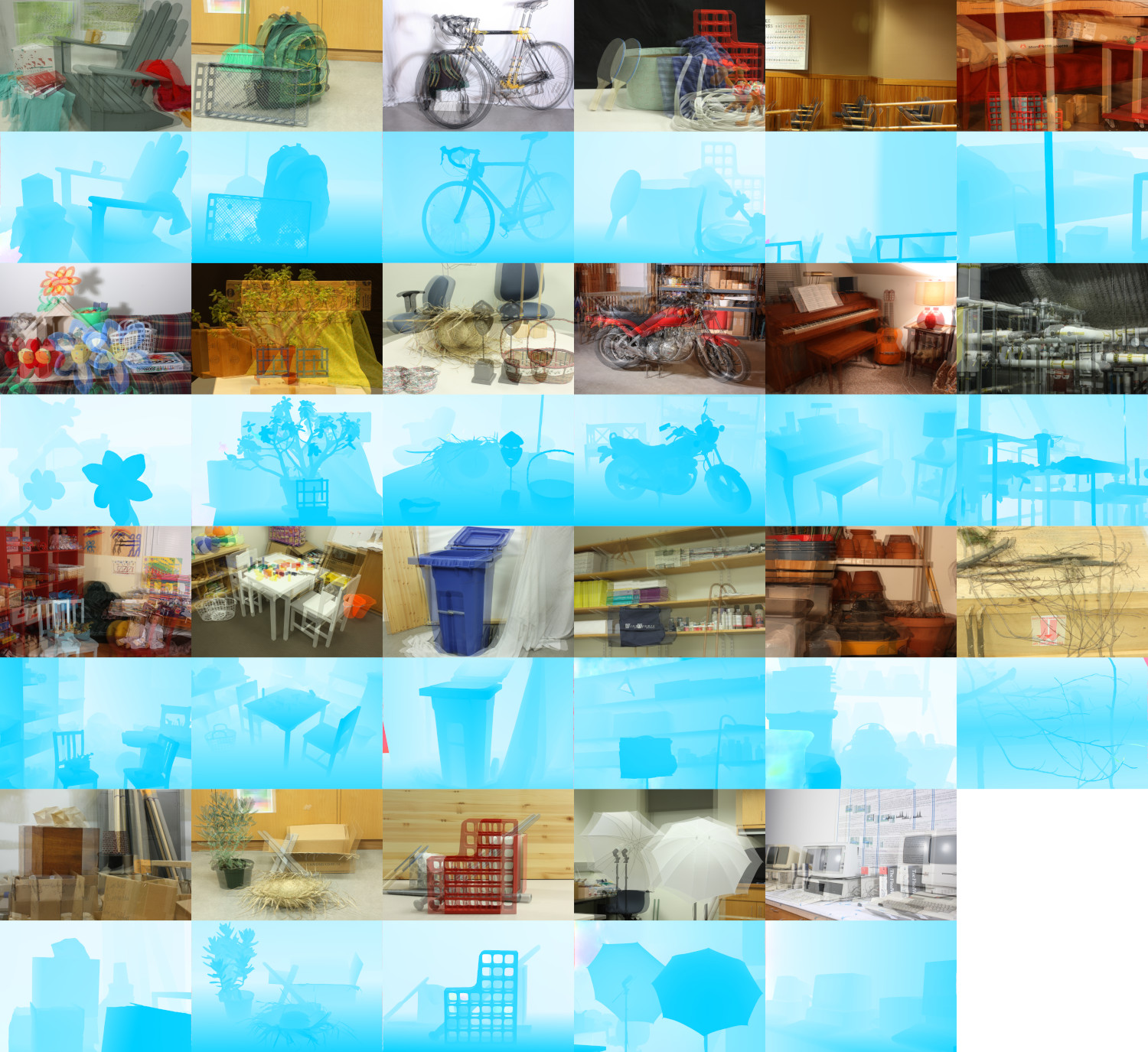}
    \caption{Qualitative results on 3K samples from the Middlebury-ST dataset~\cite{Scharstein2014Highresolutionstereo}.}
    \label{fig:sup_mbury_qualitative}
\end{figure*}

\begin{figure*}
    \centering
    \begin{subfigure}{0.45\linewidth}
        \centering
        \includegraphics[width=\linewidth]{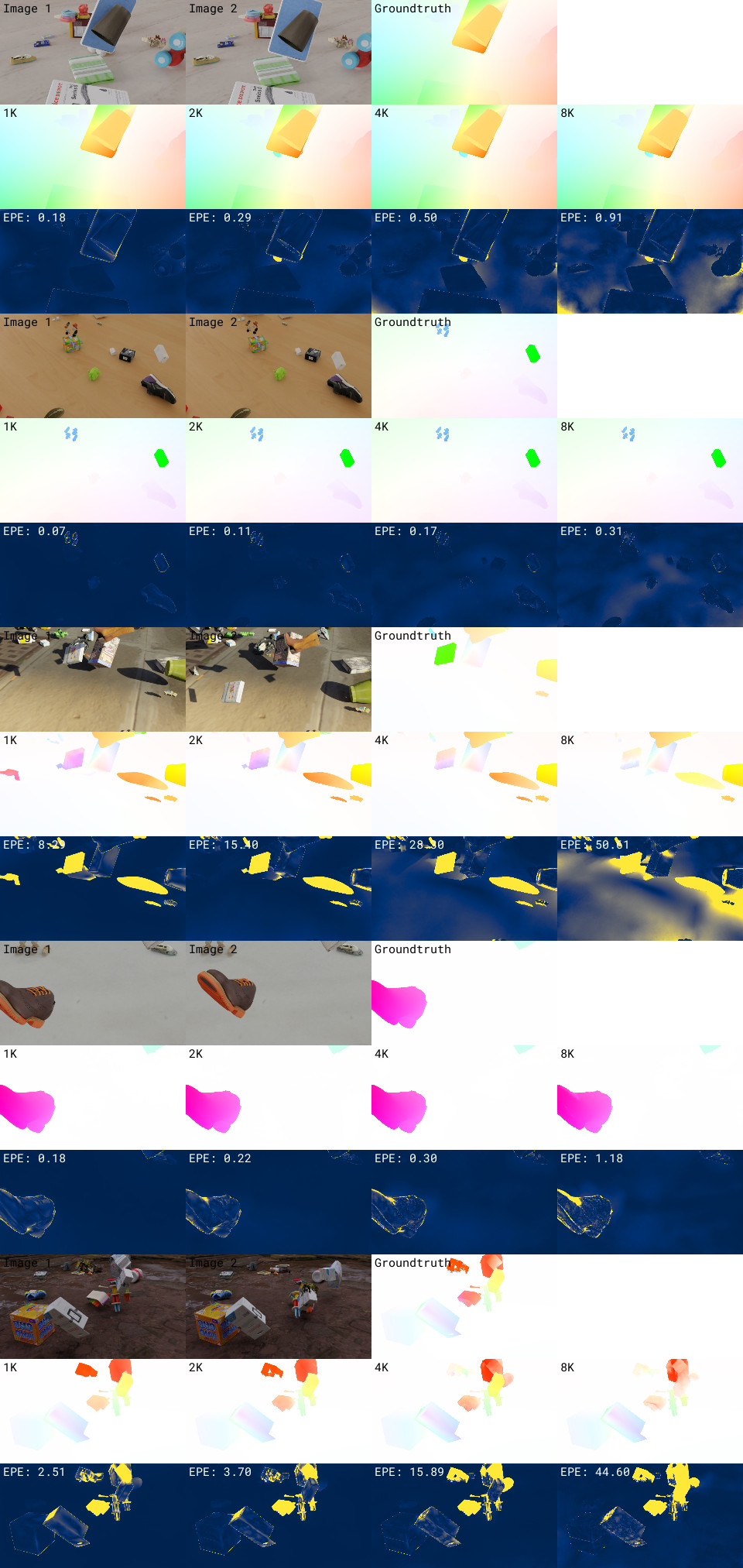}
    \end{subfigure}
    \quad
    \begin{subfigure}{0.45\linewidth}
        \centering
        \includegraphics[width=\linewidth]{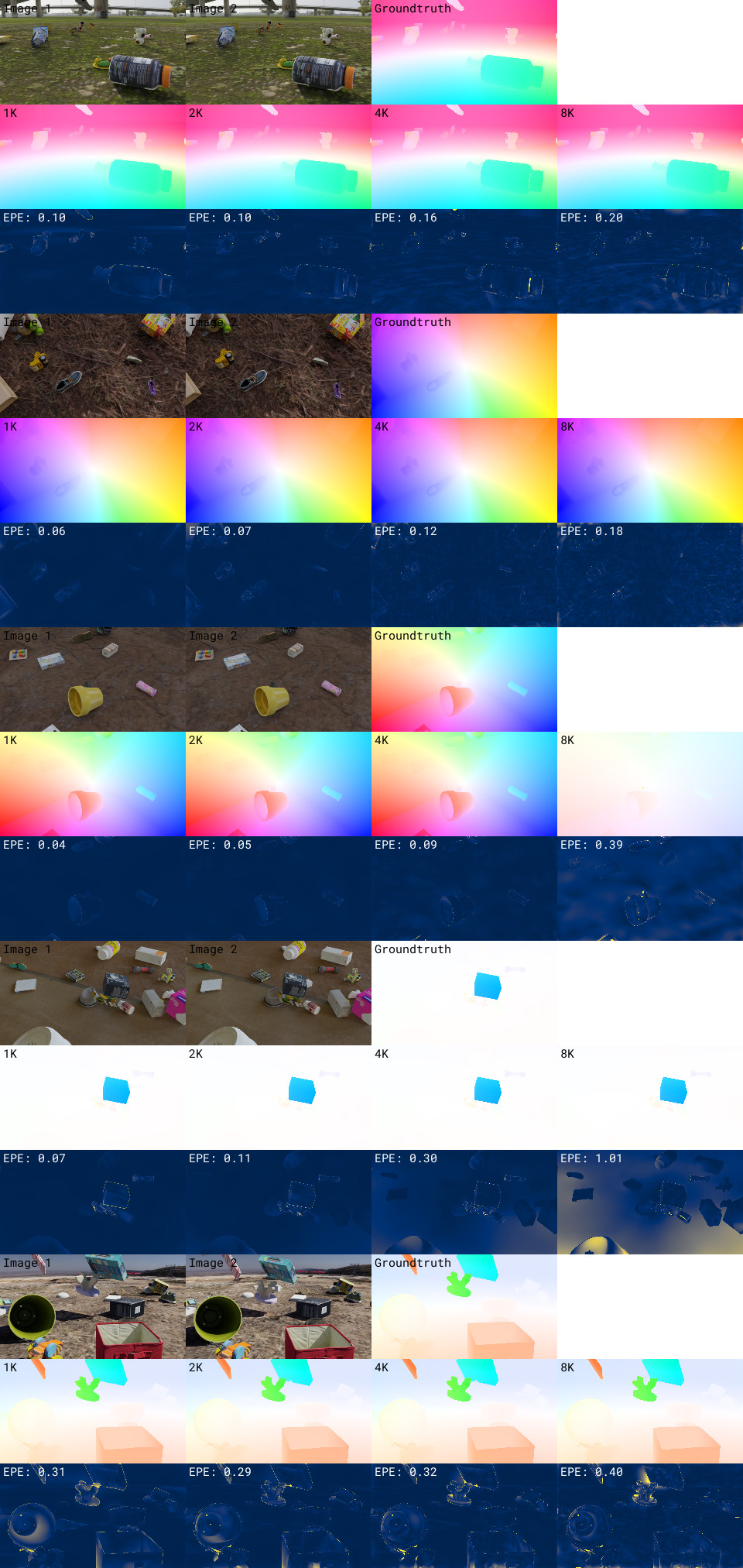}
    \end{subfigure}
    \caption{Qualitative results on 1K to 8K samples from our Kubric-NK dataset.}
    \label{fig:sup_qualitative_kubric_nk_group1-2}
\end{figure*}

\begin{figure*}
    \centering
    \begin{subfigure}{0.45\linewidth}
        \centering
        \includegraphics[width=\linewidth]{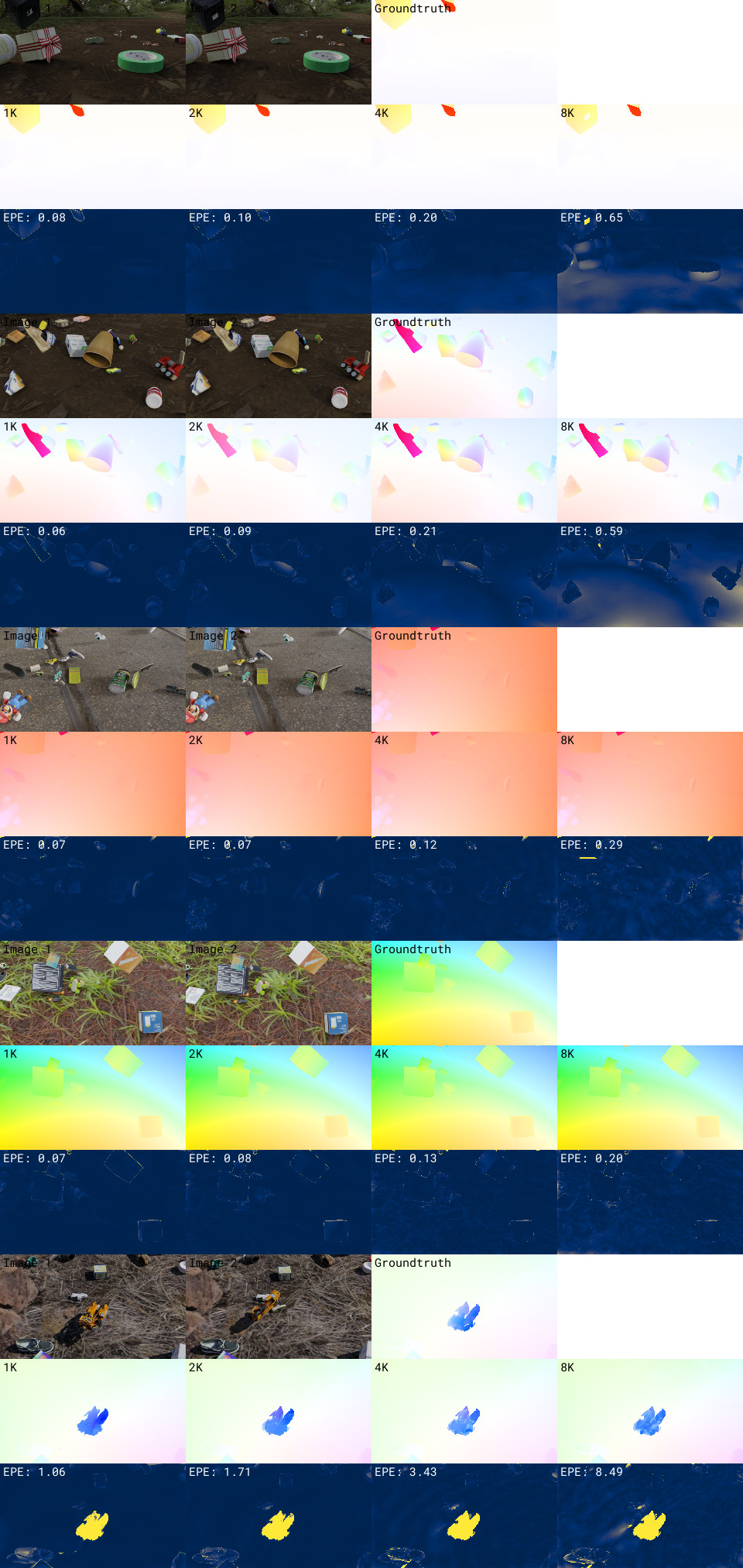}
    \end{subfigure}
    \quad
    \begin{subfigure}{0.45\linewidth}
        \centering
        \includegraphics[width=\linewidth]{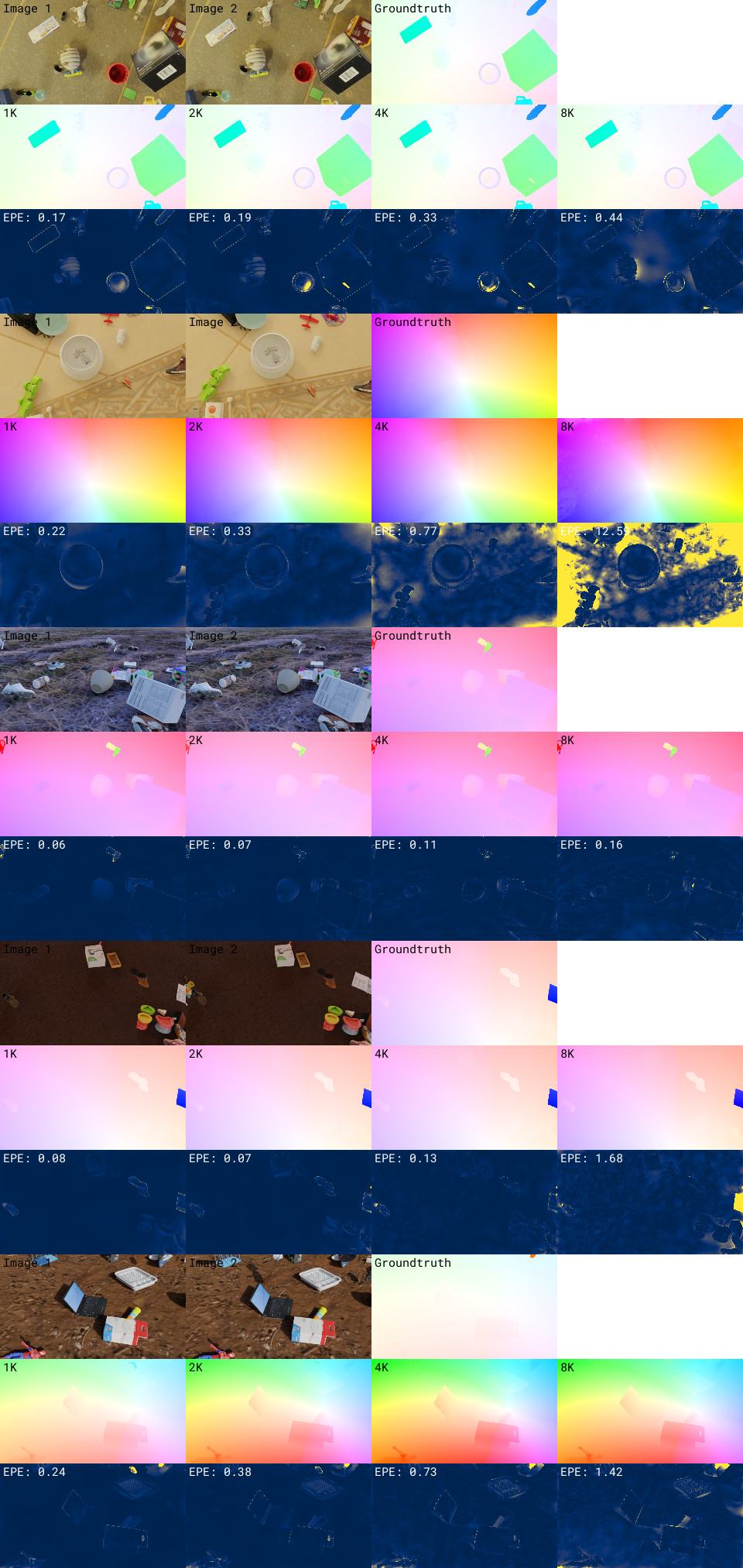}
    \end{subfigure}
    \caption{Qualitative results on 1K to 8K samples from our Kubric-NK dataset.}
    \label{fig:sup_qualitative_kubric_nk_group3-4}
\end{figure*}

\begin{figure*}
    \centering
    \begin{subfigure}{0.45\linewidth}
        \centering
        \includegraphics[width=\linewidth]{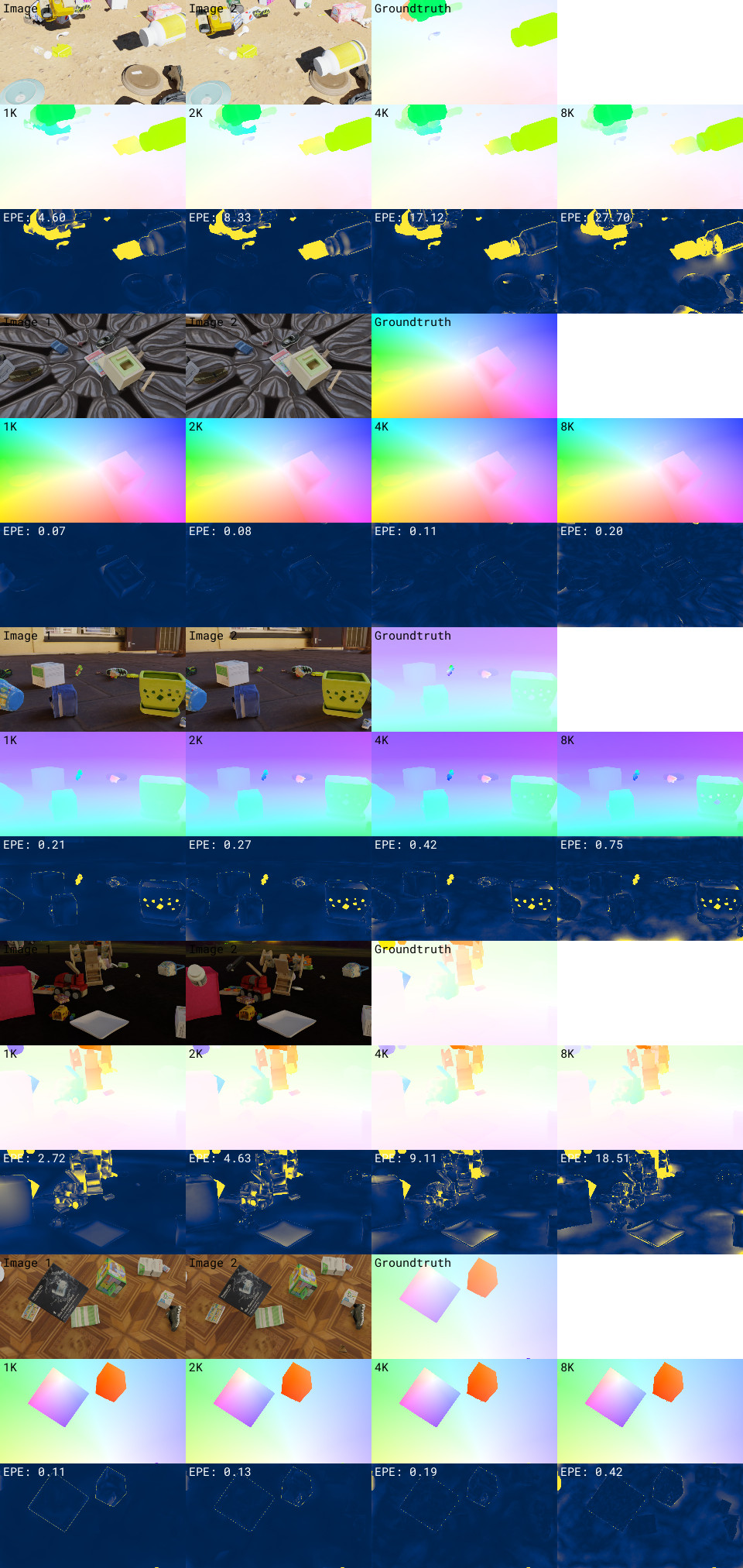}
    \end{subfigure}
    \quad
    \begin{subfigure}{0.45\linewidth}
        \centering
        \includegraphics[width=\linewidth]{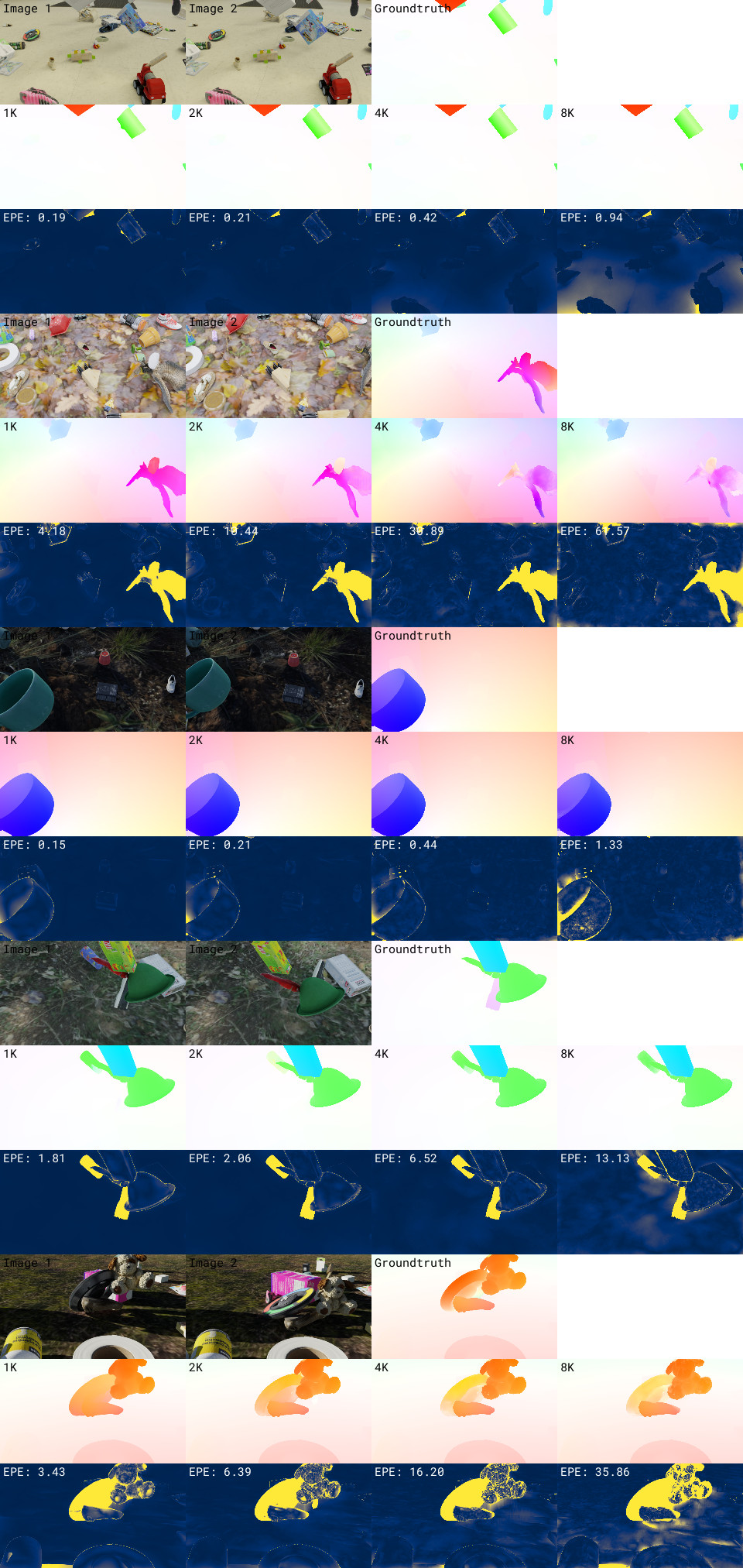}
    \end{subfigure}
    \caption{Qualitative results on 1K to 8K samples from our Kubric-NK dataset.}
    \label{fig:sup_qualitative_kubric_nk_group5-6}
\end{figure*}

This section shows some additional qualitative results of DPFlow on some high-resolution inputs.
We demonstrate results on samples at 4K resolution from DAVIS~\cite{PontTuset20172017DAVISChallenge} in~\cref{fig:sup_davis_qualitative}, 3K from Middlebury-ST~\cite{Scharstein2014Highresolutionstereo} in~\cref{fig:sup_mbury_qualitative}, and 1K to 8K from Kubric-NK in Figs.~\ref{fig:sup_qualitative_kubric_nk_group1-2}, \ref{fig:sup_qualitative_kubric_nk_group3-4}, \ref{fig:sup_qualitative_kubric_nk_group5-6}.

\end{document}